\documentclass[11pt]{article}

\usepackage[preprint]{acl}
\usepackage{times}
\usepackage{latexsym}

\usepackage{algorithm}
\usepackage[noend]{algpseudocode}  % or without [noend] if you prefer

\usepackage[T1]{fontenc}
\usepackage[utf8]{inputenc}
\usepackage {booktabs}
\usepackage{microtype}

\usepackage{inconsolata}

\usepackage{graphicx}
\usepackage{subfigure}
\usepackage{bm}
\usepackage{dblfloatfix}

\usepackage{xcolor,colortbl}
\usepackage{booktabs}
\usepackage{multirow}
\usepackage{array}
\usepackage{siunitx}

\usepackage[most]{tcolorbox}
\usepackage{xcolor}

\definecolor{PTGreen}{RGB}{92,134,110}
\definecolor{PTGreenDark}{RGB}{52,92,75}
\definecolor{PTLightGreen}{RGB}{242,250,245}
\definecolor{PTWhite}{RGB}{255,255,255}

\definecolor{Gray}{gray}{0.9}

\usepackage{amsmath}
\usepackage{amssymb}
\usepackage{amsthm}
\usepackage{soul}

\usepackage{cleveref}
\crefname{figure}{Figure}{Figures}

\usepackage{authblk}

\usepackage{xspace}
\newcommand{\method}{W2T\xspace}
\newcommand{\methodfull}{\textbf{W2T} (\textbf{W}eight-\textbf{to}-\textbf{T}oken)\xspace}
\newcommand{\maintablesize}{\footnotesize}
\theoremstyle{plain}
\newtheorem{theorem}{Theorem}[section]
\newtheorem{proposition}[theorem]{Proposition}

\theoremstyle{definition}
\newtheorem{definition}[theorem]{Definition}

\theoremstyle{remark}

\NewDocumentCommand{\zehong}
{ mO{} }{\textcolor{cyan}{\textsuperscript{\textit{Zehong}}\textsf{\textbf{\small[#1]}}}}

\title{W2T: LoRA Weights Already Know What They Can Do}

\author{
    Xiaolong Han\textsuperscript{1} \quad
    Ferrante Neri\textsuperscript{1*} \quad
    Zijian Jiang\textsuperscript{1} \quad
    Fang Wu\textsuperscript{2} \quad
    Yanfang Ye\textsuperscript{3} \quad
    Lu Yin\textsuperscript{1} \quad
    Zehong Wang\textsuperscript{3*} \\
    \textsuperscript{1} University of Surrey \quad \textsuperscript{2} Stanford University \quad \textsuperscript{3} University of Notre Dame \\
    \textsuperscript{*}Correspondence: \texttt{f.neri@surrey.ac.uk, zwang43@nd.edu}
}

\begin{document}

\maketitle

% \begin{figure*}[!t]
%     \centering
%     \includegraphics[width=\linewidth]{figure/example.pdf}
%     \caption{}
%     \label{fig:example}
% \end{figure*}

% \input{table/example}

% \cite{example}

\begin{abstract}
Each LoRA checkpoint compactly stores task-specific updates in low-rank weight matrices, offering an efficient way to adapt large language models to new tasks and domains.
In principle, these weights already encode what the adapter does and how well it performs.
In this paper, we ask whether this information can be read directly from the weights, without running the base model or accessing training data.
A key obstacle is that a single LoRA update can be factorized in infinitely many ways.
Without resolving this ambiguity, models trained on the factors may fit the particular factorization rather than the underlying update.
To this end, we propose \methodfull, which maps each LoRA update to a provably canonical form via QR decomposition followed by SVD, so that all equivalent factorizations share the same representation.
The resulting components are then tokenized and processed by a Transformer to produce a weight-space embedding.
Across language and vision LoRA collections, W2T achieves strong results on attribute classification, performance prediction, and adapter retrieval, demonstrating that LoRA weights reliably indicate model behavior once factorization ambiguity is removed.
Code is available \href{https://github.com/xiaolonghan2000/Weight2Token}{here.}
\end{abstract}
\begin{figure}[!t]
    \centering
    \includegraphics[width=1\linewidth, trim={1cm 1cm 1cm 0.5cm}, clip]{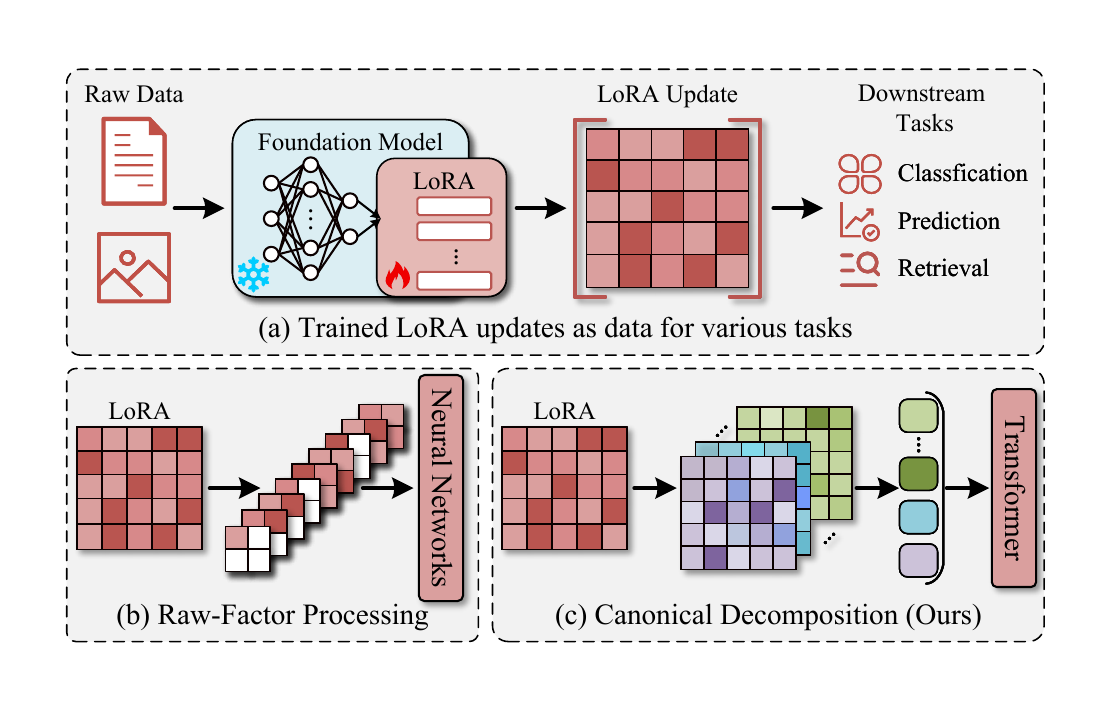}
    \caption{
    % \zehong{1. Don't say it "Motivation for W2T". It should be a better one. 2. In (b), (c), (d), don't use "artifacts", use "lora" directly. 3. in (b,c,d), don't use the term "modeling". }
    \textbf{LoRA updates already encode downstream behaviors, but reading them requires a symmetry-aware representation.}
    (a) LoRA checkpoints trained from raw data are treated as data objects for downstream tasks.
    (b) Symmetry-agnostic processing of raw LoRA factors.
    (c) Symmetry-aware decomposition of LoRA updates into canonical components.
    }
    \label{fig:motivation}
\end{figure}
\section{Introduction}
Large language models (LLMs) have shown strong performance across a wide range of domains \citep{wang2024gft,wang2026reasoning}.
To specialize them for downstream tasks, additional adaptation is often needed.
Low-Rank Adaptation (LoRA) has become a popular and efficient solution \citep{hu2022lora,dettmers2023qlora,liu2024dora}.
Each LoRA checkpoint stores task-specific updates in a pair of low-rank matrices $(\mathbf{B}, \mathbf{A})$, making it compact and easy to save, share, and reuse.
This has led to rapidly growing collections of LoRA checkpoints
% , both in large-scale serving systems and in reusable repositories 
\citep{sheng2023s,gabrielsson2025compress,huanglorahub}.
In principle, the weights should already contain information about what the LoRA does and how well it performs.
The question is whether we can read it.

However, reading adapter weights is far from straightforward.
Many shared LoRA checkpoints come with incomplete metadata, limited training details, or no accessible evaluation setup \citep{horwitz2025learning}.
In such cases, the weight file itself may be the only reliable source of information.
The main alternative is to probe each checkpoint through LLM inference, but this is expensive at scale and sensitive to prompt design \citep{zhuo2024prosa,wang2025beyond,wang2025generative}.
These constraints motivate a weight-space perspective:
\emph{Can we infer the attributes of a LoRA checkpoint directly from its weights alone, without raw training data or LLM inference?}
A positive answer would enable scalable indexing, search, capability discovery, and performance estimation for large LoRA collections \citep{putterman2025gl,schurholt2024towards,kahana2025can,salama2024dataset,horwitzwe2025we,horwitz2025learning}.

Making such weight-only inference reliable is difficult because LoRA weights do not have a unique factorization.
For a LoRA update $\Delta \mathbf{W} = \mathbf{B}\mathbf{A}$, any invertible matrix $\mathbf{G}$ gives an equivalent factor pair $(\mathbf{B}\mathbf{G},\, \mathbf{G}^{-1}\mathbf{A})$ with the same update matrix.
% This $\mathrm{GL}(r)$ reparameterization symmetry means that equivalent LoRA checkpoints can look very different in raw factors, so raw factor similarity is not a reliable indicator of update similarity.
This $\mathrm{GL}(r)$ reparameterization symmetry means that the same LoRA update $\Delta \mathbf{W}$ can be represented by many different factor pairs $(\mathbf{B}, \mathbf{A})$.
% As a result, methods that operate directly on the factor matrices, for example by flattening them, may learn coordinate-dependent signals rather than stable update structure.
As a result, methods that operate directly on the factor pair $(\mathbf{B}, \mathbf{A})$, for example by flattening them, may fail to learn the true similarity between LoRAs, since two very different-looking factor pairs can correspond to the same $\Delta \mathbf{W}$.

Existing methods for learning from LoRA weights mainly follow two directions.
One direction applies generic encoders such as MLPs, CNNs, or ViTs directly to raw or reshaped LoRA factors, without accounting for the underlying symmetry \citep{unterthiner2020predicting,schurholt2021self,jin2024conditional,wang2025scaling,shahroz2025oral}.
Another direction \citep{putterman2025gl} handles $\mathrm{GL}(r)$ symmetry in the predictor architecture through dedicated equivariant layers.
Despite this progress, a practical solution should simultaneously satisfy three requirements: resolving the non-identifiability induced by $\mathrm{GL}(r)$ symmetry, preserving the structure of each LoRA update, and remaining compatible with standard scalable architectures.
Existing methods address these requirements only partially, either learning from non-identifiable raw factors or relying on specialized equivariant architectures without canonicalizing the input parameterization itself.

% In this work, we propose \methodfull to tackle the above challenges.
% % that first resolves LoRA symmetry and then models the resulting decomposed components as structured tokens.
% W2T applies QR decomposition followed by singular value decomposition (QR--SVD) to each LoRA update, yielding a $\mathrm{GL}(r)$-invariant representation: a set of rank-wise components, each characterized by an output direction, an input direction, and a singular value.
% These rank-wise components are then converted into rank tokens through modulated tokenization.
% The resulting tokens are finally composed hierarchically within each position and across positions to produce a global weight space embedding.
% The key insight is not that we inject new information, but that we unlock what the weights already encode by providing the right representational lens.

In this work, we propose \methodfull, a framework that meets all three requirements.
The key idea is to resolve the LoRA symmetry before downstream modeling.
Rather than learning directly from arbitrary factor pairs $(\mathbf{B}, \mathbf{A})$, W2T maps all factorizations of the same LoRA update $\Delta \mathbf{W}$ to a shared canonical decomposition.
Concretely, W2T applies QR decomposition followed by singular value decomposition (SVD) to each LoRA update, producing a structured set of rank-wise components, each defined by an output direction, an input direction, and a singular value.
This canonicalization removes factorization ambiguity at the data level, so that equivalent LoRA parameterizations are aligned before they are processed by the model.
Built on top of this symmetry-resolved representation, W2T converts the components into tokens and uses a Transformer to yield a LoRA embedding for diverse downstream applications, including attribute prediction, behavior prediction, and retrieval.

% Our main contributions are as follows:
% \begin{itemize}
% \item We introduce a QR--SVD-based canonicalization procedure for LoRA updates that yields a $\mathrm{GL}(r)$-invariant representation and resolves reparameterization non-identifiability at the representation level.
% \item We introduce a token-based modeling framework that converts canonical rank-wise components into structured tokens and composes them hierarchically within and across positions.
% \item Through extensive experiments on language and vision LoRA collections, we show that resolving reparameterization symmetry at the representation level enables strong weight-only prediction of adapter attributes, performance, and cross-task similarity.
% \end{itemize}

Our main contributions are as follows:
\begin{itemize}
\item We identify LoRA factorization non-uniqueness as a core obstacle to LoRA weight-space learning and resolve it with a canonical representation of each LoRA.
\item We propose W2T, which canonically tokenizes LoRA updates and models them with a Transformer in weight space.
\item We validate W2T on diverse language and vision LoRA collections, showing strong results on attribute prediction, performance estimation, and similarity modeling.
\end{itemize}

\section{Related Work}

\paragraph{Low-rank adaptation.}
LoRA models task-specific updates with a low-rank factorization, \(\Delta \mathbf{W} = \mathbf{B}\mathbf{A}\), enabling parameter-efficient adaptation of large language models \citep{hu2022lora}.
Subsequent work has largely focused on improving LoRA as a fine-tuning method, from adaptation quality and efficiency to practical deployment.
Some work targets training efficiency: QLoRA reduces the memory cost of fine-tuning through quantization \citep{dettmers2023qlora}, while methods such as LoRA-FA \citep{zhang2023lora} and LoRA+ \citep{hayou2024lora+} further reduce memory usage or improve optimization through factor-specific update schemes.
Other work targets adaptation quality and parameterization: AdaLoRA reallocates rank budget across modules \citep{zhangadaptive}, PiSSA initializes adapters from principal singular components of pretrained weights \citep{meng2024pissa}, and DoRA decomposes magnitude and direction to better match full fine-tuning \citep{liu2024dora}.
More recent methods further explore pruning, expert allocation, and mixture-style low-rank adaptation \citep{zhou2025lora,gao2025mola,fan2025make,sun2025stronger}.

A related development shifts LoRA beyond its role as a fine-tuning method toward a reusable checkpoint format.
Some work studies how to serve large collections of LoRA adapters efficiently \citep{sheng2023s,gabrielsson2025compress}, while other work studies how separately trained LoRAs can be reused or combined across tasks and concepts \citep{huanglorahub,gu2023mix}.
Recent analyses also identify shared spectral structure across large collections of LoRA checkpoints, providing further evidence that LoRA weights themselves carry meaningful structure \citep{kaushik2025universal}.
Together, these developments make LoRA checkpoints abundant, lightweight, and increasingly reusable, making it natural to ask what can be learned from the LoRA weights themselves.

\paragraph{Weight space learning.}
Weight Space Learning (WSL) treats model checkpoints as input data and learns representations directly in the weight space.
Early studies showed that weights encode informative signals for downstream behavior, including model accuracy and class-level properties \citep{unterthiner2020predicting,eilertsen2020classifying}.
Subsequent work developed more advanced WSL models, including symmetry-aware architectures for deep weight spaces and neural functionals \citep{navon2023equivariant,zhou2023permutation,zhou2024universal,schurholt2024towards,wang2026molecular}.
In prior WSL on standard architectures, symmetry handling is mostly permutation-centric, whereas LoRA introduces a different, continuous $\mathrm{GL}(r)$ reparameterization symmetry.

For LoRA-specific weight modeling, recent work spans statistics-based inference, coordinate-space retrieval/editing, and conditional parameter generation \citep{salama2024dataset,liu2024lora,dravid2024interpreting,jin2024conditional,wang2025scaling,shahroz2025oral,liang2025drag}.
These studies show that LoRA checkpoints carry rich signals, yet most existing pipelines still rely on raw coordinates, leaving $\mathrm{GL}(r)$ reparameterization ambiguity largely unresolved.
GLNet handles this issue in the predictor architecture through specialized equivariant layers and nonlinear modules \citep{putterman2025gl}.
However, it carries symmetry handling throughout the predictor and therefore places strict requirements on the model architecture.
% \zehong{This statement should be better. We shouldn't state W2T is another approach for respecting the LoRA symmetry. We should say that explicitly modeling (W2T) the symmetry is a better solution than implicitly modeling the symmetry (GLNet). But we should carefully think about the benefits. }
Taken together, these directions still leave an important gap.
Some methods learn directly from non-identifiable raw factors, while symmetry-aware approaches rely on specialized equivariant architectures.
What remains lacking is a solution that resolves the ambiguity induced by $\mathrm{GL}(r)$ symmetry, preserves the internal structure of each LoRA update, and stays compatible with standard scalable architectures.
% \method is built for this setting by canonicalizing each LoRA update before downstream learning.
\method addresses this gap. 
\begin{figure*}[!t]
    \centering
    \includegraphics[width=1\linewidth, trim={1cm 1cm 1cm 0.5cm}, clip]{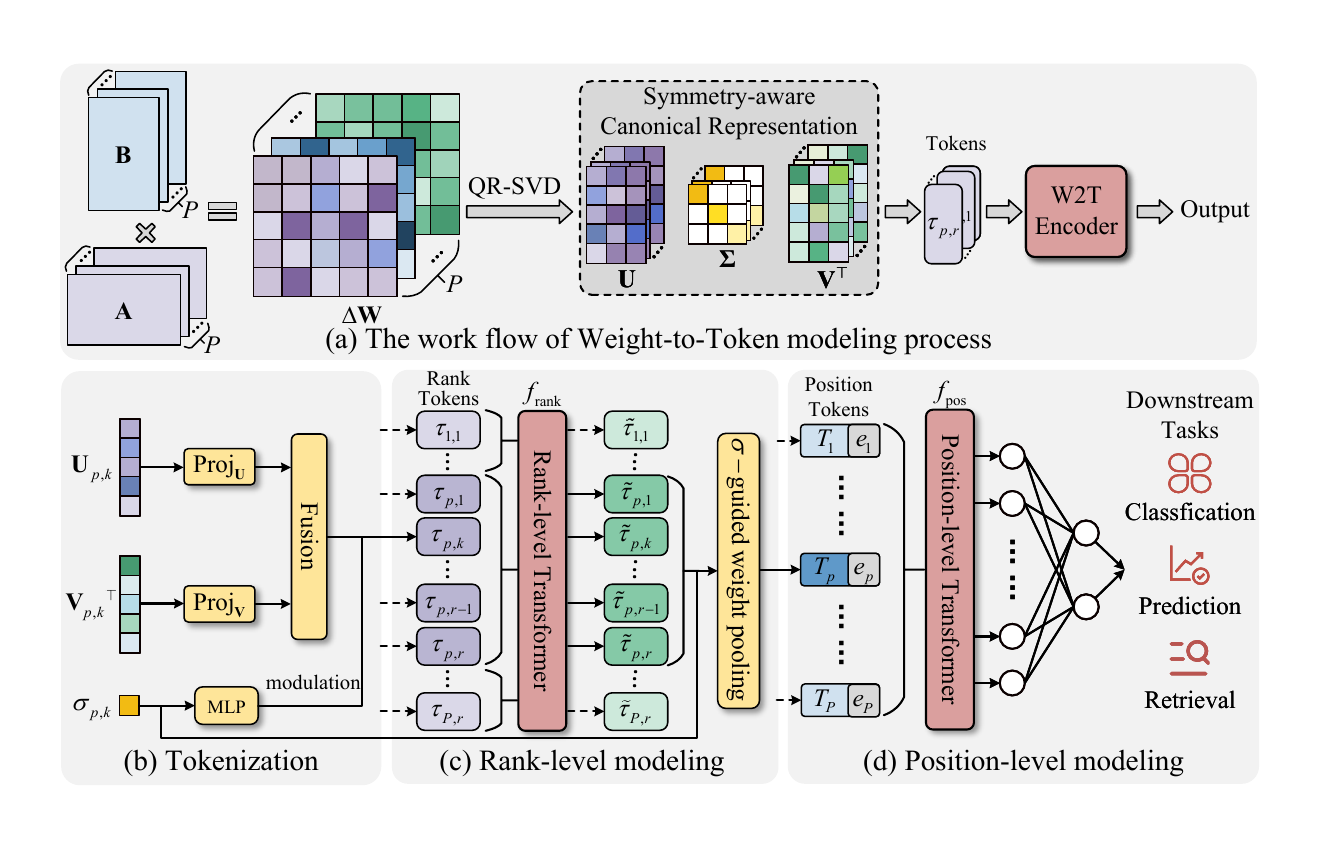}
    \caption{\textbf{Overview of the proposed \method modeling framework for LoRA weight space.}
    (a) LoRA factors are transformed into a canonical representation via symmetry-aware QR--SVD decomposition, yielding rank-wise components under a fixed deterministic convention.
    (b) Each rank-wise component is projected into a rank token through modulated tokenization.
    (c) Rank tokens within each position are seen as a set and aggregated via $\sigma$-guided pooling.
    (d) Position tokens are globally modeled to produce a weight-space embedding for downstream tasks.}
    \label{fig:framework}
\end{figure*}

% \section{How to Identify the Canonical Object of a LoRA Update?}
\section{Identifying Canonical Object of LoRA}
\label{sec:symmetry}

A LoRA update is parameterized by a factor pair
$\mathbf{B} \in \mathbb{R}^{d_{\text{out}}\times r}$ and
$\mathbf{A} \in \mathbb{R}^{r\times d_{\text{in}}}$,
with induced matrix update $\Delta \mathbf{W} = \mathbf{B}\mathbf{A}$.
Once LoRA checkpoints are treated as learning objects, the central question is no longer how to process a particular factor pair, but what should represent the update itself.
This question arises because different factorizations can induce the same $\Delta \mathbf{W}$, so raw factors depend on parameterization rather than on the update.
We therefore seek a canonical decomposition of $\Delta \mathbf{W}$ that assigns every LoRA update a unique, deterministic representation, regardless of which factor pair $(\mathbf{B},\mathbf{A})$ produced it.

\paragraph{Canonical object.}
Since $\mathrm{rank}(\Delta \mathbf{W})\le r$, the update admits an exact $r$-slot SVD,
\begin{equation}
\Delta \mathbf{W}
= \mathbf{U}_r \boldsymbol{\Sigma}_r \mathbf{V}_r^\top
= \sum_{k=1}^{r}\sigma_k \mathbf{u}_k \mathbf{v}_k^\top,
\label{eq:compact_svd}
\end{equation}
with trailing zero components when $\mathrm{rank}(\Delta \mathbf{W})<r$.
% We take this rank-wise singular decomposition to be the canonical object of the LoRA update.
% It depends only on $\Delta W$, not on a particular factor pair.
% Hence equivalent factorizations induce the same singular values and singular subspaces, up to the standard non-uniqueness of SVD.
We take this rank-wise singular decomposition as the canonical object of the LoRA update: it depends only on $\Delta \mathbf{W}$, so equivalent factorizations yield the same singular values and singular subspaces, up to the standard sign and degeneracy ambiguities of SVD.

% \begin{proposition}[Canonical representation under LoRA reparameterization]
% \label{prop:repr_invariance}
% Let $(B,A)$ and $(B',A')$ satisfy
% \[
% (B',A')=(BG,\,G^{-1}A)
% \quad\text{for some } G\in\mathrm{GL}(r).
% \]
% Then they induce the same update matrix $\Delta W$, and therefore the same decomposition in Eq.~\eqref{eq:compact_svd}, up to the intrinsic sign and degeneracy ambiguities of SVD.
% Under a fixed deterministic convention, the representation is identical.
% \end{proposition}

% Proposition~\ref{prop:repr_invariance} shows that the canonical object of a LoRA update is not the raw factor pair, but the rank-wise singular decomposition of the induced update matrix.
\begin{proposition}[$\mathrm{GL}(r)$-invariance]
\label{prop:repr_invariance}
If $(\mathbf{B}',\mathbf{A}')=(\mathbf{B}\mathbf{G},\,\mathbf{G}^{-1}\mathbf{A})$ for some $\mathbf{G}\in\mathrm{GL}(r)$, then $\mathbf{B}'\mathbf{A}'=\mathbf{B}\mathbf{A}$, and the decomposition in Eq.~\eqref{eq:compact_svd} is identical under a fixed sign and ordering convention.
\end{proposition}

That is, regardless of which factor pair produces a given $\Delta \mathbf{W}$, the canonical decomposition is the same.

% \paragraph{Construction from low-rank factors.}
% The remaining question is how to recover this canonical object without forming the full dense matrix $\Delta W$.
% Let
% \[
% B=Q_B R_B,\qquad A^\top=Q_A R_A,
% \]
% be thin QR decompositions.
% Then
% \[
% \Delta W = BA = Q_B(R_B R_A^\top)Q_A^\top,
% \]
% so the singular structure of $\Delta W$ is determined by the core matrix
% \[
% M = R_B R_A^\top \in \mathbb{R}^{r\times r}.
% \]
% If
% \[
% M=\hat U \Sigma \hat V^\top,
% \]
% then
% \[
% U=Q_B\hat U,\qquad V=Q_A\hat V
% \]
% gives an exact SVD of $\Delta W$.

% \begin{proposition}[QR-based construction]
% \label{prop:qr_svd}
% The SVD of the core matrix $M=R_B R_A^\top$ induces exactly the same singular values and singular subspaces as direct SVD on $\Delta W=BA$.
% Therefore, the canonical representation in Eq.~\eqref{eq:compact_svd} can be recovered directly from the low-rank factors.
% \end{proposition}

% Proposition~\ref{prop:qr_svd} completes the construction: the canonical object is not only well defined, but also directly computable from low-rank LoRA factors.

\paragraph{Construction from low-rank factors.}
In practice, only the low-rank factors $(\mathbf{B},\mathbf{A})$ are stored; explicitly forming $\Delta \mathbf{W}\in\mathbb{R}^{d_{\text{out}}\times d_{\text{in}}}$ is prohibitive.
We show the canonical decomposition can be obtained directly from $(\mathbf{B},\mathbf{A})$ via thin QR factorizations and a small SVD.
Let $\mathbf{B}=\mathbf{Q}_B \mathbf{R}_B$ and $\mathbf{A}^\top=\mathbf{Q}_A \mathbf{R}_A$ be thin QR decompositions. Then
\begin{equation}
    \Delta \mathbf{W} = \mathbf{B}\mathbf{A} = \mathbf{Q}_B\underbrace{(\mathbf{R}_B \mathbf{R}_A^\top)}_{\mathbf{M}}\mathbf{Q}_A^\top,
\end{equation}
where $\mathbf{M}\in\mathbb{R}^{r\times r}$ is a small core matrix that captures the entire singular structure of $\Delta \mathbf{W}$.
Computing $\mathbf{M}=\hat{\mathbf{U}}\boldsymbol{\Sigma}\hat{\mathbf{V}}^\top$ and setting $\mathbf{U}=\mathbf{Q}_B\hat{\mathbf{U}}$, $\mathbf{V}=\mathbf{Q}_A\hat{\mathbf{V}}$ gives an exact SVD of $\Delta \mathbf{W}$.

\begin{proposition}[QR-based construction]
\label{prop:qr_svd}
The SVD of $\mathbf{M}=\mathbf{R}_B \mathbf{R}_A^\top$ yields exactly the same singular values and singular subspaces as direct SVD on $\Delta \mathbf{W}=\mathbf{B}\mathbf{A}$.
\end{proposition}

Proposition~\ref{prop:qr_svd} ensures the canonical decomposition is fully recoverable from the stored factors, with cost $\mathcal{O}((d_{\text{out}}+d_{\text{in}})r^2+r^3)$ instead of $\mathcal{O}(d_{\text{out}}\,d_{\text{in}}\,r)$.
\section{W2T: Weight-to-Token Modeling}
\label{sec:w2t}

Section~\ref{sec:symmetry} establishes that the canonical object of a LoRA update is its rank-wise singular decomposition $\{(\mathbf{u}_{p,k},\mathbf{v}_{p,k},\sigma_{p,k})\}_{k=1}^{r}$, recoverable directly from the stored factors.
We now describe W2T, which turns this canonical representation into a learnable embedding for downstream tasks.

\paragraph{Framework overview.}
As illustrated in Figure~\ref{fig:framework}(a), W2T takes raw LoRA factors $(\mathbf{B},\mathbf{A})$ at every adapted weight matrix $p$ (e.g., $\mathbf{W}_Q$ in layer~3), applies the canonicalized decomposition from Section~\ref{sec:symmetry} to obtain the symmetry-aware canonical representation $(\mathbf{U},\boldsymbol{\Sigma},\mathbf{V}^\top)$, and converts each rank-wise component into a token $\boldsymbol{\tau}_{p,k}$.
These tokens are then processed by the W2T encoder, which operates in two hierarchical stages: a rank-level Transformer models interactions among rank tokens within each position and aggregates them into position tokens via $\sigma$-guided pooling; a position-level Transformer then models cross-position dependencies and produces a global embedding for downstream classification, prediction, and retrieval.

\paragraph{Tokenization.}
Each rank component is mapped to a fixed-length token (Figure~\ref{fig:framework}(b)).
Because $\mathbf{u}_{p,k}$ and $\mathbf{v}_{p,k}$ lie in different spaces (output and input, respectively), we encode them with separate projectors and fuse:
\begin{equation}
\mathbf{z}_{p,k}
=
\mathbf{W}_{\mathrm{fuse}}
\big[
\phi_u(\mathbf{u}_{p,k}) \,\|\, \phi_v(\mathbf{v}_{p,k})
\big],
\end{equation}
where $\phi_u$, $\phi_v$ are learnable projectors and $\|$ denotes concatenation.
We then inject the singular value $\sigma_{p,k}$ via modulation.
We first apply a stabilizing transform $\hat{\sigma}_{p,k}=\log(1+\sigma_{p,k})$, then map it through an MLP to obtain scale and shift parameters $({\gamma}_{p,k},{\beta}_{p,k}) = \mathrm{MLP}_\sigma(\hat{\sigma}_{p,k})$:
\begin{equation}
\boldsymbol{\tau}_{p,k}
=
\mathbf{z}_{p,k}\odot\big(1+\tanh({\gamma}_{p,k})\big)+{\beta}_{p,k},
\end{equation}
where $\odot$ is element-wise multiplication.
The modulation lets $\sigma_{p,k}$ control how strongly each rank component is expressed in the token, without altering the directional content encoded in $\mathbf{z}_{p,k}$.

\paragraph{Rank-level modeling.}
Recall that each position $p$ corresponds to one LoRA-adapted weight matrix, yielding $r$ rank tokens $\boldsymbol{\tau}_{p,1:r}$.
A rank-level Transformer $f_{\mathrm{rank}}$ models interactions among these $r$ tokens within the same position (Figure~\ref{fig:framework}(c)).
The enriched tokens are then aggregated into a single position token via $\sigma$-guided weighted pooling:
\begin{equation}
\mathbf{t}_p
=
\mathrm{RankPool}\!\left(f_{\mathrm{rank}}(\boldsymbol{\tau}_{p,1:r}),\,\sigma_{p,1:r}\right),
\end{equation}
% \zehong{What is RankPool. We need a definition.}
where \(\mathrm{RankPool}\) computes a weighted sum of rank tokens using the normalised singular values as weights, so that dominant components contribute more strongly to the position representation.

\paragraph{Position-level modeling.}
Position tokens are augmented with learnable embeddings for layer depth and module type (Figure~\ref{fig:framework}(d)):
\begin{equation}
\tilde{\mathbf{t}}_p
=
\mathbf{t}_p
+
\mathbf{e}_{\mathrm{layer}}(\ell(p))
+
\mathbf{e}_{\mathrm{module}}(m(p)),
\end{equation}
then processed by a position-level Transformer $f_{\mathrm{pos}}$ that captures cross-position dependencies:
\begin{equation}
\mathbf{g}_{1:|\mathcal{P}|}
=
f_{\mathrm{pos}}(\tilde{\mathbf{t}}_{1:|\mathcal{P}|}).
\end{equation}
Finally, attention pooling yields a global embedding $\mathbf{h} = \sum_{p\in\mathcal{P}} \alpha_p \mathbf{g}_p$, which serves as the LoRA checkpoint representation for downstream tasks.

\section{Experiments}
\label{sec:experiments}

\begin{table*}[!t]
    \caption{Attribute classification results (\%). The best value in each column is highlighted in \textbf{bold}.}
    \label{tab:att_class}
        \centering
        \maintablesize
        \setlength{\tabcolsep}{3.2pt}
        \renewcommand{\arraystretch}{1.05}
    \begin{tabular}{
            l
            S[table-format=1.2]
            S[table-format=2.2]
            S[table-format=2.2]
            S[table-format=2.2]
            S[table-format=2.2]
            S[table-format=2.2]
            S[table-format=2.2]
            S[table-format=2.2]
            S[table-format=2.2]
        }
            \toprule
            \sisetup{
                input-ignore = {\,\textbf},
                table-number-alignment = right,
                table-text-alignment = right,
                table-align-text-before = false
            }
            \multirow{2}{*}{Algorithm}
            & \multicolumn{3}{c}{GoEmotions-LoRA}
            & \multicolumn{3}{c}{CelebA-LoRA}
            & \multicolumn{3}{c}{CUB-LoRA} \\
            \cmidrule(lr){2-4}\cmidrule(lr){5-7}\cmidrule(lr){8-10}
            & \multicolumn{1}{c}{Macro-F1$\uparrow$} & \multicolumn{1}{c}{Micro-F1$\uparrow$} & \multicolumn{1}{c}{AUROC$\uparrow$}
            & \multicolumn{1}{c}{Macro-F1$\uparrow$} & \multicolumn{1}{c}{Micro-F1$\uparrow$} & \multicolumn{1}{c}{AUROC$\uparrow$}
            & \multicolumn{1}{c}{Macro-F1$\uparrow$} & \multicolumn{1}{c}{Micro-F1$\uparrow$} & \multicolumn{1}{c}{AUROC$\uparrow$} \\
            \midrule
            MLP   & 2.87 & 21.67 & 49.40 & 7.73 & 40.64 & 51.60 & 7.44 & 26.23 & 47.81 \\
            CNN   & 0.00 & 0.00 & 51.78 & 15.61 & 50.15 & 60.66 & 1.40 & 18.48 & 50.22 \\
            ViT   & 1.16 & 14.52 & 49.78 & 6.07 & 37.32 & 49.11 & 1.21 & 16.83 & 50.41 \\
            GLNet & 1.17 & 9.97 & 47.77 & 48.61 & 74.83 & 89.53 & 10.73 & 38.69 & 69.84 \\
            \rowcolor{black!8}\method (Ours)
                  & \textbf{5.67} & \textbf{24.84} & \textbf{61.25}
                  & \textbf{50.38} & \textbf{75.02} & \textbf{89.64}
                  & \textbf{18.34} & \textbf{43.24} & \textbf{73.72} \\
            \bottomrule
        \end{tabular}
\end{table*}

\subsection{Setup}

\paragraph{Experimental Setting.}
Our experiments target the setting where the only input is a complete LoRA checkpoint. Each checkpoint is treated as one sample in a weight dataset, and supervision is attached at the checkpoint level. Concretely, the label of a LoRA sample denotes either the semantic attributes inherited from the data used to train that adapter, or the downstream evaluation scores measured from that adapter.

\paragraph{LoRA datasets and tasks.}
Our main supervised collections span vision and language, with 8:1:1 train/val/test splits over LoRA samples. On the vision side, LoRAs are trained on Stable Diffusion v1.4 \citep{rombach2022high} from CelebA \citep{liu2015deep} and CUB \citep{wah2011caltech}, yielding CelebA-LoRA and CUB-LoRA. On the language side, all LoRAs use Llama-3.2-3B \citep{grattafiori2024llama}. GoEmotions-LoRA \citep{demszky2020goemotions} supports semantic characterization, while ARC-LoRA supports performance prediction on ARC-Easy. Separately, for adapter retrieval, we build a mixed language adapter pool from four task datasets: ARC-Challenge \citep{clark2018think}, BoolQ \citep{clark2019boolq}, GSM8K \citep{cobbe2021training}, and MBPP \citep{austin2021program}.
Detailed checkpoint-level label definitions, generation protocols, split rules, and retrieval pool construction are deferred to Appendix \ref{sec:appendix_data_prep}.

\paragraph{Baselines.}
We compare W2T with one symmetry-aware baseline and several symmetry-agnostic raw-factor encoders.
GLNet \citep{putterman2025gl} is specifically designed to respect the intrinsic $\mathrm{GL}(r)$ symmetry of LoRA factorization.
MLP, CNN \citep{jin2024conditional,liang2025drag}, and ViT \citep{wang2025scaling,shahroz2025oral} instead encode flattened or reshaped raw LoRA factors without explicit symmetry modeling.
For generic baselines, we compare their encoder architectures rather than their original end-to-end application setups. The details of all baselines and their implementation in our setting are described in Appendix \ref{sec:impl_details}.

\paragraph{Evaluation metrics.}
For attribute classification, we report macro-F1, micro-F1, and area under the ROC curve (AUROC). For performance prediction, we report mean absolute error (MAE), root mean squared error (RMSE), Pearson correlation, and Spearman correlation. For adapter retrieval, we report normalized discounted cumulative gain at 10 (NDCG@10).
% These metrics respectively evaluate overall classification quality, numerical error and ranking consistency, and retrieval quality.
All reported results are averaged over five independent runs.

% ============================================================
% RQ1: What can we decode from LoRA weights alone?
% ============================================================
\subsection{What Can We Decode from LoRA Weights Alone?}

We test this question along three axes: attribute classification, performance prediction, and adapter retrieval, corresponding to adapter semantics, downstream quality, and task-level similarity.

\paragraph{Attribute classification.}\label{sec:attr_cls}

Given a LoRA adapter trained on a subset of data defined by certain attributes (e.g., facial features in CelebA, bird species traits in CUB, or emotion categories in GoEmotions), we train a linear classifier to predict those attributes solely from the adapter's weight representation.
As shown in Table \ref{tab:att_class}, on CelebA-LoRA and CUB-LoRA, several baselines already obtain non-trivial scores, confirming that LoRA weights do encode semantic information about the training data.
GoEmotions-LoRA is more challenging and exposes larger gaps between representations.
Raw-factor baselines already recover some semantic information, but \method achieves the strongest overall results and the most reliable performance across datasets.
These results suggest that the main challenge is not whether semantic information exists in LoRA weights, but how effectively a representation can decode it.

\paragraph{Performance prediction.}\label{sec:perf_pred}
\begin{table}[!t]
    \centering
    \caption{Performance prediction results (\%) on ARC-LoRA. Best values are in \textbf{bold}.}
    \label{tab:perf_pred}
    \maintablesize
    \setlength{\tabcolsep}{2.2pt}
    \renewcommand{\arraystretch}{1.05}
    \begin{tabular}{
        ll
        S[table-format=1.2]
        S[table-format=1.2]
        S[table-format=2.2]
        S[table-format=2.2]
    }
        \toprule
        Method & Split & \multicolumn{1}{c}{MAE$\downarrow$} & \multicolumn{1}{c}{RMSE$\downarrow$} & \multicolumn{1}{c}{Pearson$\uparrow$} & \multicolumn{1}{c}{Spearman$\uparrow$} \\
        \midrule
        \multirow{2}{*}{MLP}
            & Valid & 1.58 & 4.75 & 68.12 & 70.49 \\
            & Test  & 1.29 & 3.31 & 71.88 & 71.55 \\
        \midrule
        \multirow{2}{*}{CNN}
            & Valid & 0.92 & 1.98 & 94.29 & 75.62 \\
            & Test  & 0.89 & 2.04 & 89.48 & 75.39 \\
        \midrule
        \multirow{2}{*}{ViT}
            & Valid & 0.80 & 2.00 & 94.23 & 70.94 \\
            & Test  & 0.77 & 1.76 & 92.17 & 71.30 \\
        \midrule
        \multirow{2}{*}{GLNet}
            & Valid & 0.63 & 3.05 & 91.05 & 94.45 \\
            & Test  & 0.43 & 1.90 & 92.78 & 94.16 \\
        \midrule
        \rowcolor{black!8}
            & Valid & {\bfseries 0.34} & {\bfseries 0.72} & {\bfseries 99.37} & {\bfseries 95.26} \\
        \rowcolor{black!8}
        \multirow{-2}{*}{\raisebox{-0.6ex}{\cellcolor{black!8}\method (Ours)}}
            & Test  & {\bfseries 0.32} & {\bfseries 0.67} & {\bfseries 98.98} & {\bfseries 95.35} \\
        \bottomrule
    \end{tabular}
\end{table}

Given a collection of LoRA adapters trained on ARC-Easy under diverse hyperparameter configurations, we train a regressor to predict each adapter's downstream accuracy solely from its weight representation.
Unlike attribute classification, where labels reflect training data identity, performance prediction requires the representation to encode quality signals that generalize across different training conditions.
As shown in Table \ref{tab:perf_pred}, LoRA weights do encode how well the adapter will perform downstream.
MAE and RMSE measure numerical fidelity, while Pearson and Spearman measure ranking consistency across checkpoints.
\method performs strongly on both error-based and rank-based metrics.
The same symmetry-aware pattern appears here: methods that resolve reparameterization ambiguity consistently outperform raw-factor baselines.
This indicates that representation quality mainly affects calibration and ordering accuracy, not only absolute error.

\paragraph{Adapter retrieval}\label{sec:fewshot_retrieval}
% \begin{table}[!t]
%     \centering
%     \caption{Few-shot query retrieval under dataset shift (\%) on a mixed LoRA pool. Best values are in \textbf{bold}.}
%     \label{tab:fewshot_retrieval}
%     \maintablesize
%     \setlength{\tabcolsep}{2.2pt}
%     \renewcommand{\arraystretch}{1.05}
%     \begin{tabular}{
%         l
%         S[table-format=2.2]
%         S[table-format=2.2]
%         S[table-format=2.2]
%         S[table-format=2.2]
%     }
%         \toprule
%         Method & \multicolumn{1}{c}{Hit@1$\uparrow$} & \multicolumn{1}{c}{Hit@10$\uparrow$} & \multicolumn{1}{c}{MRR$\uparrow$} & \multicolumn{1}{c}{NDCG@10$\uparrow$} \\
%         \midrule
%         MLP & 19.35 & 75.18 & 35.33 & 22.22 \\
%         CNN & 13.32 & 80.79 & 31.01 & 15.76 \\
%         ViT & 25.95 & {\bfseries 99.44} & 48.16 & 20.22 \\
%         GLNet & 45.44 & 78.68 & 57.09 & 43.00 \\
%         \rowcolor{black!8}\method (Ours) & {\bfseries 53.30} & 92.57 & {\bfseries 66.11} & {\bfseries 52.37} \\
%         \bottomrule
%     \end{tabular}
% \end{table}

\begin{table}[!t]
    \centering
    \caption{Adapter retrieval results (NDCG@10, \%).
    All learned encoders are trained on ARC-Easy-LoRA performance prediction and transferred to retrieval without retrieval-specific tuning.
    RawCos is a training-free baseline that ranks adapters by cosine similarity in raw LoRA weight space. Best values are in \textbf{bold}.}
    \label{tab:fewshot_retrieval}
    \maintablesize
    \setlength{\tabcolsep}{2.0pt}
    \renewcommand{\arraystretch}{1.05}
    \begin{tabular}{
        l
        S[table-format=2.2]
        S[table-format=2.2]
        S[table-format=2.2]
        S[table-format=2.2]
        S[table-format=2.2]
        S[table-format=2.2]
    }
    \toprule
    Method & \multicolumn{1}{c}{Average} & \multicolumn{1}{c}{ARC-C} & \multicolumn{1}{c}{BoolQ} & \multicolumn{1}{c}{GSM8K} & \multicolumn{1}{c}{MBPP} \\
    \midrule
    RawCos & 35.73 & 39.90 & 45.17 & \textbf{33.41} & 24.42 \\
    \midrule
    MLP & 27.82 & 35.61 & 0.00 & 0.00 & 75.65 \\
    CNN & 24.54 & 17.56 & 42.80 & 4.84 & 32.95 \\
    ViT & 25.33 & 13.32 & 33.15 & 29.55 & 25.30 \\
    GLNet & 43.97 & 77.84 & 0.00 & 16.65 & 81.39 \\
    \rowcolor{black!8}\method (Ours) & \textbf{65.71} & \textbf{99.14} & \textbf{50.76} & 24.97 & \textbf{87.96} \\
    \bottomrule
    \end{tabular}
\end{table}

Given a mixed pool of LoRA adapters spanning multiple downstream tasks, we ask whether weight-space representations can identify adapters associated with a target task using only weight-space similarity, without access to task labels or adapter outputs.
This tests a transfer dimension absent from the previous two tasks: the encoder is trained on a different source task and must generalize to unseen task categories at retrieval time.
Table \ref{tab:fewshot_retrieval} evaluates this setting: a weight-space encoder trained on one source task is used to retrieve adapters associated with related downstream tasks in a mixed pool.
All learned encoders are trained only for ARC-Easy-LoRA performance prediction, then frozen and transferred to retrieval on a mixed adapter pool (ARC-Challenge, BoolQ, GSM8K, MBPP) without retrieval-specific tuning.
We also include a training-free baseline, RawCos, which directly ranks by cosine similarity in raw LoRA weight space.
\method achieves the best average retrieval quality and is strongest on three of four datasets, indicating that canonical token representations transfer well beyond the source task.
On GSM8K, RawCos is higher, which is consistent with target-shift effects: under a frozen transfer backbone trained on another task, distribution mismatch can hurt task-specific ranking, while raw similarity may remain competitive on outlier domains.
Even with this outlier, \method remains the best overall method by average NDCG@10.

\begin{figure}[t]
    \centering
    \includegraphics[width=1\linewidth]{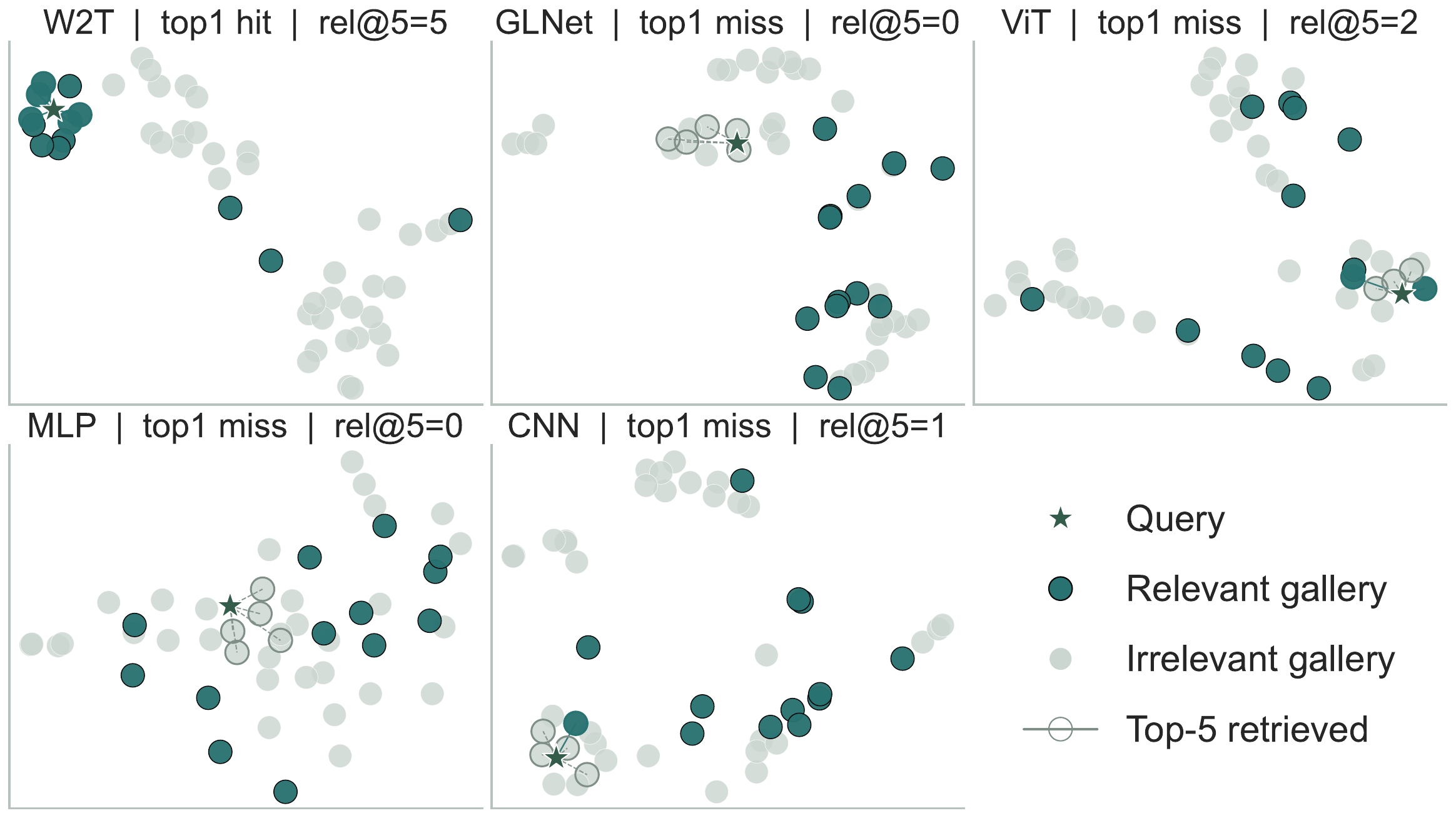}
    \caption{Local retrieval neighborhood for a 8-shot BoolQ query. Each subplot shows the same query (star) amid a shared pool. W2T places relevant adapters (dark) closest to the query with top-1 hit and rel@5=5, while baselines surface more irrelevant neighbors.}
    \label{fig:fewshot_query_scatter}
\end{figure}

Figure \ref{fig:fewshot_query_scatter} illustrates this advantage from a local geometric perspective. For a fixed BoolQ query adapter, we visualize the shared gallery neighborhood retrieved across all models. \method concentrates relevant adapters near the query, whereas baselines produce mixed neighborhoods with more irrelevant top-ranked results.
Appendix \ref{app:retrieval_cluster_additional} further analyzes this transfer from the cost--quality trade-off of adapter retrieval.

\paragraph{Ablation study.}
\begin{figure}[t]
    \centering
    \includegraphics[width=1\linewidth]{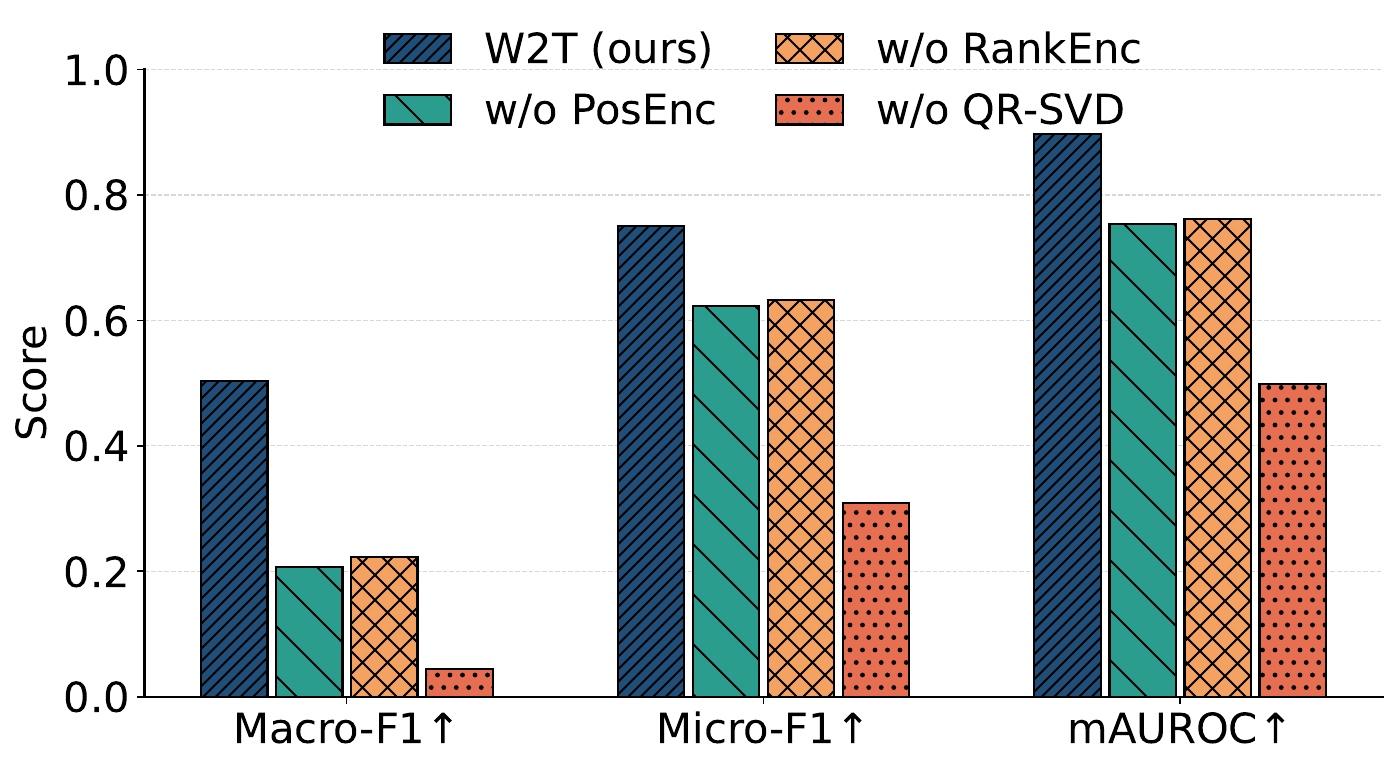}
    \caption{Ablation on attribute classification. Results are reported on CelebA-LoRA. Removing QR--SVD causes the largest performance drop, while removing rank-level or position-level modeling leads to smaller but consistent degradations.}
    \label{fig:ablation}
\end{figure}
Figure \ref{fig:ablation} gives a clear pattern.
When rank-level modeling or position-level modeling is removed, performance drops in a consistent but limited way, which means both parts are useful.
The much bigger change appears when canonicalized tokenization (QR--SVD) is removed and the raw $\mathbf{B}$ and $\mathbf{A}$ factors are instead flattened and concatenated as the input representation: metrics fall sharply and become close to raw-factor baselines.
These results indicate that symmetry-aware canonicalization is the primary driver of W2T's advantage over raw-factor baselines.
% while rank-level and position-level modeling each provide additional but incremental gains on top of a well-posed input representation.}

\paragraph{Summary.}
Across classification, performance prediction, and retrieval, these results show that LoRA weights do contain usable signals about adapter semantics, downstream quality, and task-level similarity.
\method reads these signals most effectively, motivating a deeper analysis of whether explicit symmetry resolution is the key factor behind these gains.

% ============================================================
% RQ2: Is symmetry resolution the critical ingredient?
% ============================================================
\subsection{Why Does Symmetry Resolution Drive the Performance Gap?}

The results above show a consistent advantage for symmetry-aware methods, but do not yet isolate \emph{why}.
We analyze this question at two levels: model behavior and decomposition fidelity.

\begin{figure}[t]
    \centering
    \includegraphics[width=1.0\linewidth]{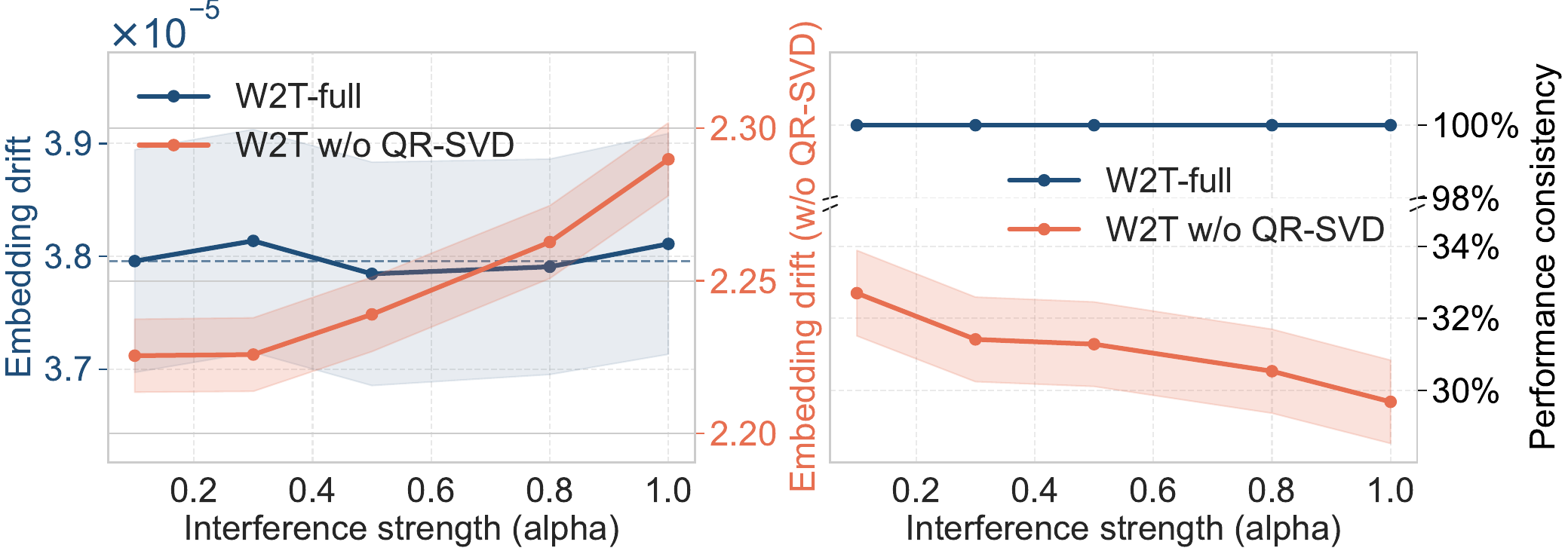}
    \caption{Practical GL($r$) invariance.
    The horizontal axis is perturbation strength $\alpha$ in random GL($r$) reparameterization (larger $\alpha$ means stronger perturbation).
    Left: embedding drift under increasing perturbation.
    Right: prediction consistency under the same perturbation schedule.}
    \label{fig:gl_invariance}
\end{figure}

\begin{table}[!t]
\centering
\caption{Practical equivalence of QR-based core SVD and direct SVD. Unless noted otherwise, values are per-pair medians; Factor cosine reports the mean of per-pair minimum subspace cosines.}
\label{tab:qrsvd_equiv}
\maintablesize
\setlength{\tabcolsep}{0pt}
\renewcommand{\arraystretch}{1.08}
\begin{tabular*}{0.98\linewidth}{@{\extracolsep{\fill}}>{\raggedright\arraybackslash}p{0.49\linewidth}>{\raggedleft\arraybackslash}p{0.33\linewidth}@{}}
\toprule
Property & Value \\
\midrule
Singular-value gap $\delta_{\sigma}\downarrow$ & $3.13\times10^{-6}$ \\
Update gap $\delta_{\Delta W}\downarrow$ & $1.38\times10^{-5}$ \\
Factor cosine $(U/V)$ (mean)$\uparrow$ & $0.999946/0.999955$ \\
Time (ms), SVD / QR--SVD & $73.17/0.779$ \\
Speedup$\uparrow$ & $93.18\times$ \\
\bottomrule
\end{tabular*}
\end{table}

\begin{figure}[t]
    \centering
    \includegraphics[width=1\linewidth]{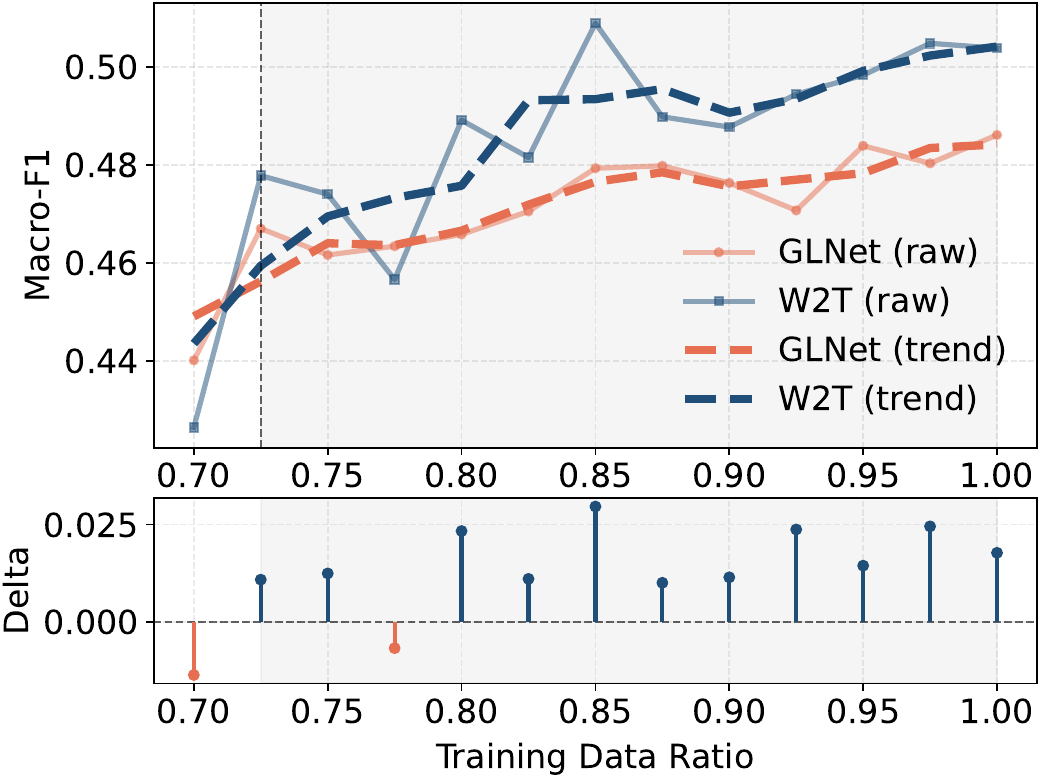}
    \caption{Scaling behavior from a training-data ratio of 0.7 onward. Markers are raw Macro-F1 values. Dashed curves are 3-point centered moving-average trends. The lower panel reports pointwise $\Delta$. Over this displayed range, W2T follows a steeper upward trend than GLNet.}
    \label{fig:scaling}
\end{figure}

\paragraph{Practical GL($r$) invariance.}
Figure \ref{fig:gl_invariance} shows that canonical decomposition stabilizes both representation and prediction under $\mathrm{GL}(r)$ reparameterization.
Concretely, for each checkpoint we apply $(B,A)\!\mapsto\!(BG,\,G^{-1}A)$ with random $G$, where $\alpha$ controls the perturbation strength.
Each point aggregates 30 random transforms at that $\alpha$ (mean with confidence interval), and $\alpha=1.0$ is the strongest setting.
With QR--SVD canonicalization, embedding drift stays near zero and prediction agreement remains high; without it, both degrade sharply.
This aligns with the ablation results in Section \ref{sec:attr_cls}, where removing QR--SVD causes by far the largest drop in downstream performance.

Table \ref{tab:qrsvd_equiv} serves as an implementation check.
Under float32 precision, reduced QR--SVD introduces only negligible numerical discrepancies relative to direct SVD while still reducing decomposition time by nearly two orders of magnitude, confirming that the efficient construction is faithful enough for practical use.
These tiny gaps mainly come from finite-precision rounding and accumulated multiplication error along different decomposition paths, rather than a mismatch in the underlying representation.
The prediction consistency in Figure \ref{fig:gl_invariance} further shows that these numerical effects do not break the effective uniqueness of our representation under GL($r$) transformations.

\paragraph{Scaling with Supervision.}
Figure \ref{fig:scaling} compares W2T and GLNet over the displayed training-data ratios, from 0.7 onward.
It shows both raw Macro-F1 points and smoothed trends, where the trend is computed with a 3-point centered moving average (with edge padding at boundaries).
We use the smoothed curves only for readability, while conclusions are drawn from the raw points.
From a training-data ratio of 0.7 onward, the smoothed trend of W2T rises more steeply than that of GLNet, indicating that it benefits more from additional supervision.
Since both methods are symmetry-aware, this pattern points to a difference in how they use supervision rather than whether they handle symmetry at all.
A likely explanation is architectural: GLNet relies on equivariant linear layers, whereas W2T uses a Transformer over canonical tokens, whose higher modeling capacity appears to make better use of additional supervision.

\begin{table}[!t]
\centering
\caption{Base-model transfer on CelebA-LoRA (\%). All models are trained on SD1.4 LoRA checkpoints and evaluated zero-shot on SD1.5 at the same rank ($r=8$). Oracle denotes the in-domain \method result on SD1.5 under the same protocol.}
\label{tab:ood}
\maintablesize
\setlength{\tabcolsep}{3.2pt}
\renewcommand{\arraystretch}{1.05}
\begin{tabular}{l S[table-format=2.2] S[table-format=2.2] S[table-format=2.2]}
\toprule
Method & \multicolumn{1}{c}{Macro-F1$\uparrow$} & \multicolumn{1}{c}{Micro-F1$\uparrow$} & \multicolumn{1}{c}{mAUROC$\uparrow$} \\
\midrule
MLP   & 11.51 & 43.91 & 52.25 \\
CNN   & 5.01 & 32.06 & 49.25 \\
ViT   & 6.08 & 37.32 & 49.70 \\
GLNet & 46.41 & 73.70 & 89.16 \\
\rowcolor{black!8}\method (Ours) & \textbf{48.06} & \textbf{73.84} & \textbf{89.23} \\
\midrule
Oracle (SD1.5) & 50.24 & 74.82 & 89.30 \\
\bottomrule
\end{tabular}
\end{table}

% ============================================================
% RQ3: Does the representation generalize?
% ============================================================

\subsection{How Well Does the Learned Representation Transfer?}

The previous experiments operate within the training-time factorization and backbone configuration.
We now test cross-backbone transfer by training on SD1.4 and evaluating zero-shot on SD1.5 at the same rank ($r=8$).
Table \ref{tab:ood} reports this base-model transfer setting together with an in-domain oracle on SD1.5.
Raw-factor baselines remain far below the symmetry-aware methods under this transfer.
\method slightly outperforms GLNet across all three metrics and remains close to the SD1.5 oracle.
This suggests that the learned representation transfers across nearby base models rather than overfitting a single backbone instance.

\section{Conclusion}

We presented \method, a framework that canonicalizes LoRA updates via QR--SVD decomposition, tokenizes the resulting rank-wise components, and models them with a Transformer.
Across attribute classification, performance prediction, adapter retrieval, and base-model transfer, \method achieves strong results over raw-factor baselines and prior symmetry-aware alternatives, with ablations confirming that canonical decomposition is the primary driver of robustness.
Our findings suggest that scalable inference from LoRA weights depends on constructing a canonical, structured representation that standard architectures can learn from effectively.

% \zehong{Use paragraphs in the limitations and ethical Considerations}

\section*{Limitations}

\paragraph{Scope of the studied object.}
W2T studies standalone LoRA checkpoints as the object of weight-space inference. The current formulation assumes that each sample corresponds to a single LoRA update attached to a fixed backbone and adapter layout. It does not yet address settings such as merged adapters, composed checkpoints, or dynamic multi-adapter workflows, where the relation between stored weights and downstream behavior can be more complex.

\paragraph{Empirical coverage.}
Our experiments focus on discriminative uses of LoRA representations, including classification, regression, and similarity-based retrieval. It remains unclear how the same canonicalized representation should be used in generative settings. In particular, if a model is trained to generate canonicalized weights rather than raw LoRA factors, the target space may no longer match the format that is most natural for storage, exchange, or direct deployment.

\paragraph{Deployment considerations.}
Our retrieval study is still a proxy setting rather than a full retrieval system. The current encoder is trained through another supervised task and then reused to compute embedding similarity. A realistic large-scale retrieval model would likely require retrieval-specific training, indexing, negative sampling, and serving design. Building such a task-native retrieval pipeline remains an important direction for future work.

\section*{Ethical Considerations}
This work is a piece of basic research on weight-space representation learning. All experiments are conducted on public datasets, public model families, and openly available LoRA-style training pipelines; we do not use private user data or domain-specific confidential data.

The main ethical risk is not in the data source itself, but in what weight-space analysis could enable. Methods that infer semantics, performance, or similarity directly from checkpoints may eventually support large-scale profiling, filtering, or attribution of third-party adapters. For this reason, such tools should be used as analytical aids rather than definitive judgments about a model's capability, intent, or safety.

\bibliography{citation}

@inproceedings{hu2022lora,
title={Lo{RA}: Low-Rank Adaptation of Large Language Models},
author={Edward J Hu and yelong shen and Phillip Wallis and Zeyuan Allen-Zhu and Yuanzhi Li and Shean Wang and Lu Wang and Weizhu Chen},
booktitle={International Conference on Learning Representations},
year={2022},
}

@inproceedings{putterman2025gl,
  title={GL Equivariant Metanetworks for Learning on Low Rank Weight Spaces},
  author={Putterman, Theo and Lim, Derek and Gelberg, Yoav and Bronstein, Michael M and Jegelka, Stefanie and Maron, Haggai},
  booktitle={The Fourth Learning on Graphs Conference},
  year={2025}
}

@article{jin2024conditional,
  title={Conditional lora parameter generation},
  author={Jin, Xiaolong and Wang, Kai and Tang, Dongwen and Zhao, Wangbo and Zhou, Yukun and Tang, Junshu and You, Yang},
  journal={arXiv preprint arXiv:2408.01415},
  year={2024}
}

@inproceedings{navon2023equivariant,
  title={Equivariant architectures for learning in deep weight spaces},
  author={Navon, Aviv and Shamsian, Aviv and Achituve, Idan and Fetaya, Ethan and Chechik, Gal and Maron, Haggai},
  booktitle={International Conference on Machine Learning},
  pages={25790--25816},
  year={2023},
  organization={PMLR}
}

@inproceedings{liu2024dora,
  title={DoRA: Weight-Decomposed Low-Rank Adaptation},
  author={Liu, Shih-Yang and Wang, Chien-Yi and Yin, Hongxu and Molchanov, Pavlo and Wang, Yu-Chiang Frank and Cheng, Kwang-Ting and Chen, Min-Hung},
  booktitle={International Conference on Machine Learning},
  pages={32100--32121},
  year={2024},
  organization={PMLR}
}

@article{sheng2023s,
  title={S-lora: Serving thousands of concurrent lora adapters},
  author={Sheng, Ying and Cao, Shiyi and Li, Dacheng and Hooper, Coleman and Lee, Nicholas and Yang, Shuo and Chou, Christopher and Zhu, Banghua and Zheng, Lianmin and Keutzer, Kurt and Gonzalez, Joseph E. and Stoica, Ion},
  journal={arXiv preprint arXiv:2311.03285},
  year={2023}
}

@inproceedings{gabrielsson2025compress,
  title={Compress then Serve: Serving Thousands of LoRA Adapters with Little Overhead},
  author={Gabrielsson, Rickard Br{\"u}el and Zhu, Jiacheng and Bhardwaj, Onkar and Choshen, Leshem and Greenewald, Kristjan and Yurochkin, Mikhail and Solomon, Justin},
  booktitle={International Conference on Machine Learning},
  pages={18062--18095},
  year={2025},
  organization={PMLR}
}

@inproceedings{huanglorahub,
  title={LoraHub: Efficient Cross-Task Generalization via Dynamic LoRA Composition},
  author={Huang, Chengsong and Liu, Qian and Lin, Bill Yuchen and Pang, Tianyu and Du, Chao and Lin, Min},
  booktitle={First Conference on Language Modeling},
  year={2024}
}

@inproceedings{zhuo2024prosa,
  title={ProSA: Assessing and understanding the prompt sensitivity of LLMs},
  author={Zhuo, Jingming and Zhang, Songyang and Fang, Xinyu and Duan, Haodong and Lin, Dahua and Chen, Kai},
  booktitle={Findings of the Association for Computational Linguistics: EMNLP 2024},
  pages={1950--1976},
  year={2024}
}

@article{schurholt2021self,
  title={Self-supervised representation learning on neural network weights for model characteristic prediction},
  author={Sch{\"u}rholt, Konstantin and Kostadinov, Dimche and Borth, Damian},
  journal={Advances in Neural Information Processing Systems},
  volume={34},
  pages={16481--16493},
  year={2021}
}

@inproceedings{shahroz2025oral,
  title={ORAL: Prompting Your Large-Scale LoRAs via Conditional Recurrent Diffusion},
  author={Shahroz, Rana and Tang, Dongwen and Li, Pingzhi and Wang, Kai and Chen, Tianlong},
  booktitle={Findings of the Association for Computational Linguistics: EMNLP 2025},
  pages={1357--1370},
  year={2025}
}

@inproceedings{wang2025scaling,
  title={Scaling Up Parameter Generation: A Recurrent Diffusion Approach},
  author={Wang, Kai and Tang, Dongwen and Zhao, Wangbo and Sch{\"u}rholt, Konstantin and Wang, Zhangyang and You, Yang},
  booktitle={Advances in neural information processing systems},
  year={2025}
}

@inproceedings{liang2025drag,
title={Drag-and-Drop {LLM}s: Zero-Shot Prompt-to-Weights},
author={Liang, Zhiyuan and Tang, Dongwen and Zhou, Yuhao and Zhao, Xuanlei and Shi, Mingjia and Zhao, Wangbo and Li, Zekai and Wang, Peihao and Sch{\"u}rholt, Konstantin and Borth, Damian and Bronstein, Michael M. and You, Yang and Wang, Zhangyang and Wang, Kai},
booktitle={Advances in neural information processing systems},
year={2025},
}

@inproceedings{fan2025make,
  title={Make LoRA Great Again: Boosting LoRA with Adaptive Singular Values and Mixture-of-Experts Optimization Alignment},
  author={Fan, Chenghao and Lu, Zhenyi and Liu, Sichen and Gu, Chengfeng and Qu, Xiaoye and Wei, Wei and Cheng, Yu},
  booktitle={International Conference on Machine Learning},
  pages={15804--15832},
  year={2025},
  organization={PMLR}
}

@article{meng2024pissa,
  title={Pissa: Principal singular values and singular vectors adaptation of large language models},
  author={Meng, Fanxu and Wang, Zhaohui and Zhang, Muhan},
  journal={Advances in Neural Information Processing Systems},
  volume={37},
  pages={121038--121072},
  year={2024}
}

@inproceedings{zhangadaptive,
  title={Adaptive Budget Allocation for Parameter-Efficient Fine-Tuning},
  author={Zhang, Qingru and Chen, Minshuo and Bukharin, Alexander and He, Pengcheng and Cheng, Yu and Chen, Weizhu and Zhao, Tuo},
  booktitle={International Conference on Learning Representations},
  year={2023}
}

@inproceedings{zhou2025lora,
  title={Lora-drop: Efficient lora parameter pruning based on output evaluation},
  author={Zhou, Hongyun and Lu, Xiangyu and Xu, Wang and Zhu, Conghui and Zhao, Tiejun and Yang, Muyun},
  booktitle={Proceedings of the 31st International Conference on Computational Linguistics},
  pages={5530--5543},
  year={2025}
}

@inproceedings{gao2025mola,
  title={MoLA: MoE LoRA with layer-wise expert allocation},
  author={Gao, Chongyang and Chen, Kezhen and Rao, Jinmeng and Liu, Ruibo and Sun, Baochen and Zhang, Yawen and Peng, Daiyi and Guo, Xiaoyuan and Subrahmanian, VS},
  booktitle={Findings of the Association for Computational Linguistics: NAACL 2025},
  pages={5097--5112},
  year={2025}
}

@article{gu2023mix,
  title={Mix-of-show: Decentralized low-rank adaptation for multi-concept customization of diffusion models},
  author={Gu, Yuchao and Wang, Xintao and Wu, Jay Zhangjie and Shi, Yujun and Chen, Yunpeng and Fan, Zihan and Xiao, Wuyou and Zhao, Rui and Chang, Shuning and Wu, Weijia and Ge, Yixiao and Shan, Ying and Shou, Mike Zheng},
  journal={Advances in Neural Information Processing Systems},
  volume={36},
  pages={15890--15902},
  year={2023}
}

@inproceedings{
kahana2025can,
title={Can this Model Also Recognize Dogs? Zero-Shot Model Search from Weights},
author={Jonathan Kahana and Or Nathan and Eliahu Horwitz and Yedid Hoshen},
booktitle={Workshop on Neural Network Weights as a New Data Modality},
year={2025},
}

@article{unterthiner2020predicting,
  title={Predicting neural network accuracy from weights},
  author={Unterthiner, Thomas and Keysers, Daniel and Gelly, Sylvain and Bousquet, Olivier and Tolstikhin, Ilya},
  journal={arXiv preprint arXiv:2002.11448},
  year={2020}
}

@inproceedings{eilertsen2020classifying,
  title={Classifying the classifier: dissecting the weight space of neural networks},
  author={Eilertsen, Gabriel and J{\"o}nsson, Daniel and Ropinski, Timo and Unger, Jonas and Ynnerman, Anders},
  booktitle={European Conference on Artificial Intelligence (ECAI 2020)},
  volume={325},
  pages={1119--1126},
  year={2020},
  organization={IOS PRESS}
}

@article{dettmers2023qlora,
  title={Qlora: Efficient finetuning of quantized llms},
  author={Dettmers, Tim and Pagnoni, Artidoro and Holtzman, Ari and Zettlemoyer, Luke},
  journal={Advances in neural information processing systems},
  volume={36},
  pages={10088--10115},
  year={2023}
}

@inproceedings{sun2025stronger,
  title={A Stronger Mixture of Low-Rank Experts for Fine-Tuning Foundation Models},
  author={Sun, Mengyang and Wang, Yihao and Feng, Tao and Zhang, Dan and Zhu, Yifan and Tang, Jie},
  booktitle={International Conference on Machine Learning},
  pages={57712--57727},
  year={2025},
  organization={PMLR}
}

@article{zhou2023permutation,
  title={Permutation equivariant neural functionals},
  author={Zhou, Allan and Yang, Kaien and Burns, Kaylee and Cardace, Adriano and Jiang, Yiding and Sokota, Samuel and Kolter, J Zico and Finn, Chelsea},
  journal={Advances in neural information processing systems},
  volume={36},
  pages={24966--24992},
  year={2023}
}

@article{zhou2024universal,
  title={Universal neural functionals},
  author={Zhou, Allan and Finn, Chelsea and Harrison, James},
  journal={Advances in neural information processing systems},
  volume={37},
  pages={104754--104775},
  year={2024}
}

@inproceedings{schurholt2024towards,
  title={Towards Scalable and Versatile Weight Space Learning},
  author={Sch{\"u}rholt, Konstantin and Mahoney, Michael W and Borth, Damian},
  booktitle={International Conference on Machine Learning},
  pages={43947--43966},
  year={2024},
  organization={PMLR}
}

@article{salama2024dataset,
  title={Dataset size recovery from lora weights},
  author={Salama, Mohammad and Kahana, Jonathan and Horwitz, Eliahu and Hoshen, Yedid},
  journal={arXiv preprint arXiv:2406.19395},
  year={2024}
}

@article{dravid2024interpreting,
  title={Interpreting the weight space of customized diffusion models},
  author={Dravid, Amil and Gandelsman, Yossi and Wang, Kuan-Chieh and Abdal, Rameen and Wetzstein, Gordon and Efros, Alexei and Aberman, Kfir},
  journal={Advances in Neural Information Processing Systems},
  volume={37},
  pages={137334--137371},
  year={2024}
}

@inproceedings{horwitzwe2025we,
  title={We Should Chart an Atlas of All the World's Models},
  author={Horwitz, Eliahu and Kurer, Nitzan and Kahana, Jonathan and Amar, Liel and Hoshen, Yedid},
  booktitle={The Thirty-Ninth Annual Conference on Neural Information Processing Systems Position Paper Track},
  year={2025},
}

@inproceedings{liu2015deep,
  title={Deep learning face attributes in the wild},
  author={Liu, Ziwei and Luo, Ping and Wang, Xiaogang and Tang, Xiaoou},
  booktitle={Proceedings of the IEEE international conference on computer vision},
  pages={3730--3738},
  year={2015}
}

@techreport{wah2011caltech,
  title={The Caltech-UCSD Birds-200-2011 Dataset},
  author={Wah, Catherine and Branson, Steve and Welinder, Peter and Perona, Pietro and Belongie, Serge},
  institution={California Institute of Technology},
  number={CNS-TR-2011-001},
  year={2011}
}

@inproceedings{demszky2020goemotions,
  title={GoEmotions: A dataset of fine-grained emotions},
  author={Demszky, Dorottya and Movshovitz-Attias, Dana and Ko, Jeongwoo and Cowen, Alan and Nemade, Gaurav and Ravi, Sujith},
  booktitle={Proceedings of the 58th annual meeting of the association for computational linguistics},
  pages={4040--4054},
  year={2020}
}

@article{clark2018think,
  title={Think you have solved question answering? try arc, the ai2 reasoning challenge},
  author={Clark, Peter and Cowhey, Isaac and Etzioni, Oren and Khot, Tushar and Sabharwal, Ashish and Schoenick, Carissa and Tafjord, Oyvind},
  journal={arXiv preprint arXiv:1803.05457},
  year={2018}
}

@inproceedings{clark2019boolq,
  title={Boolq: Exploring the surprising difficulty of natural yes/no questions},
  author={Clark, Christopher and Lee, Kenton and Chang, Ming-Wei and Kwiatkowski, Tom and Collins, Michael and Toutanova, Kristina},
  booktitle={Proceedings of the 2019 conference of the north American chapter of the association for computational linguistics: Human language technologies, volume 1 (long and short papers)},
  pages={2924--2936},
  year={2019}
}

@article{cobbe2021training,
  title={Training verifiers to solve math word problems},
  author={Cobbe, Karl and Kosaraju, Vineet and Bavarian, Mohammad and Chen, Mark and Jun, Heewoo and Kaiser, Lukasz and Plappert, Matthias and Tworek, Jerry and Hilton, Jacob and Nakano, Reiichiro and others},
  journal={arXiv preprint arXiv:2110.14168},
  year={2021}
}

@article{austin2021program,
  title={Program synthesis with large language models},
  author={Austin, Jacob and Odena, Augustus and Nye, Maxwell and Bosma, Maarten and Michalewski, Henryk and Dohan, David and Jiang, Ellen and Cai, Carrie and Terry, Michael and Le, Quoc and others},
  journal={arXiv preprint arXiv:2108.07732},
  year={2021}
}

@inproceedings{rombach2022high,
  title={High-resolution image synthesis with latent diffusion models},
  author={Rombach, Robin and Blattmann, Andreas and Lorenz, Dominik and Esser, Patrick and Ommer, Bj{\"o}rn},
  booktitle={Proceedings of the IEEE/CVF conference on computer vision and pattern recognition},
  pages={10684--10695},
  year={2022}
}

@article{grattafiori2024llama,
  title={The llama 3 herd of models},
  author={Grattafiori, Aaron and Dubey, Abhimanyu and Jauhri, Abhinav and Pandey, Abhinav and Kadian, Abhishek and Al-Dahle, Ahmad and Letman, Aiesha and Mathur, Akhil and Schelten, Alan and Vaughan, Alex and others},
  journal={arXiv preprint arXiv:2407.21783},
  year={2024}
}

@inproceedings{horwitz2025learning,
  title={Learning on model weights using tree experts},
  author={Horwitz, Eliahu and Cavia, Bar and Kahana, Jonathan and Hoshen, Yedid},
  booktitle={Proceedings of the Computer Vision and Pattern Recognition Conference},
  pages={20468--20478},
  year={2025}
}

@article{liu2024lora,
  title={A lora is worth a thousand pictures},
  author={Liu, Chenxi and Takikawa, Towaki and Jacobson, Alec},
  journal={arXiv preprint arXiv:2412.12048},
  year={2024}
}

@article{zhang2023lora,
  title={Lora-fa: Memory-efficient low-rank adaptation for large language models fine-tuning},
  author={Zhang, Longteng and Zhang, Lin and Shi, Shaohuai and Chu, Xiaowen and Li, Bo},
  journal={arXiv preprint arXiv:2308.03303},
  year={2023}
}

@inproceedings{hayou2024lora+,
  title={LoRA+: Efficient Low Rank Adaptation of Large Models},
  author={Hayou, Soufiane and Ghosh, Nikhil and Yu, Bin},
  booktitle={International Conference on Machine Learning},
  year={2024},
}

@article{kaushik2025universal,
  title={The universal weight subspace hypothesis},
  author={Kaushik, Prakhar and Chaudhari, Shravan and Vaidya, Ankit and Chellappa, Rama and Yuille, Alan},
  journal={arXiv preprint arXiv:2512.05117},
  year={2025}
}

@article{wang2024gft,
  title={Gft: Graph foundation model with transferable tree vocabulary},
  author={Wang, Zehong and Zhang, Zheyuan and Chawla, Nitesh V and Zhang, Chuxu and Ye, Yanfang},
  journal={Advances in neural information processing systems},
  volume={37},
  pages={107403--107443},
  year={2024}
}

@article{wang2026molecular,
  title={Molecular Representations in Implicit Functional Space via Hyper-Networks},
  author={Wang, Zehong and Han, Xiaolong and Yang, Qi and Tang, Xiangru and Wu, Fang and Guo, Xiaoguang and Sun, Weixiang and Ma, Tianyi and Lio, Pietro and Cong, Le and Wang, Sheng and Zhang, Chuxu and Ye, Yanfang},
  journal={arXiv preprint arXiv:2601.22327},
  year={2026}
}

@article{wang2026reasoning,
  title={Why Reasoning Fails to Plan: A Planning-Centric Analysis of Long-Horizon Decision Making in LLM Agents},
  author={Wang, Zehong and Wu, Fang and Wang, Hongru and Tang, Xiangru and Li, Bolian and Yin, Zhenfei and Ma, Yijun and Li, Yiyang and Sun, Weixiang and Chen, Xiusi and others},
  journal={arXiv preprint arXiv:2601.22311},
  year={2026}
}

@inproceedings{wang2025beyond,
  title={Beyond Message Passing: Neural Graph Pattern Machine},
  author={Wang, Zehong and Zhang, Zheyuan and Ma, Tianyi and Chawla, Nitesh V and Zhang, Chuxu and Ye, Yanfang},
  booktitle={International Conference on Machine Learning},
  year={2025}
}

@inproceedings{wang2025generative,
  title={Generative Graph Pattern Machine},
  author={Wang, Zehong and Zhang, Zheyuan and Ma, Tianyi and Zhang, Chuxu and Ye, Yanfang},
  booktitle={Advances in Neural Information Processing Systems},
  year={2025}
}

\appendix
\section{Theoretical Proofs}
\label{sec:appendix_proofs}

This appendix provides formal proofs for the two theoretical statements used in Section~\ref{sec:symmetry}: (i) Proposition~\ref{prop:repr_invariance} on the invariance of the canonical rank-wise decomposition under LoRA reparameterisation, and (ii) Proposition~\ref{prop:qr_svd} on the QR-based construction of the same canonical object from low-rank factors.

\subsection{Invariance of the Canonical Object under LoRA Reparameterisation}
\label{sec:svd_inv_prove}

\paragraph{Setup.}
For a LoRA update $\Delta \mathbf{W} \in \mathbb{R}^{d_{\text{out}}\times d_{\text{in}}}$, LoRA parameterises
\begin{equation}
\Delta \mathbf{W} = \mathbf{B}\mathbf{A},
\qquad
\mathbf{B} \in \mathbb{R}^{d_{\text{out}}\times r},\;
\mathbf{A} \in \mathbb{R}^{r\times d_{\text{in}}}.
\end{equation}

\begin{definition}[LoRA Reparameterisation Symmetry]
\label{def:lora_symm_appendix}
Let $\mathrm{GL}(r)$ denote the group of invertible $r\times r$ matrices.
For any $\mathbf{G}\in \mathrm{GL}(r)$, define the transformation
\begin{equation}
(\mathbf{B},\mathbf{A}) \ \mapsto\ (\mathbf{B}',\mathbf{A}') := (\mathbf{B}\mathbf{G},\; \mathbf{G}^{-1}\mathbf{A}).
\end{equation}
\end{definition}

\paragraph{Canonical rank-wise decomposition.}
Let the SVD of $\Delta \mathbf{W}$ be
\begin{equation}
\Delta \mathbf{W} = \mathbf{U}\boldsymbol{\Sigma}\mathbf{V}^\top,
\end{equation}
where $\mathbf{U}\in\mathbb{R}^{d_{\text{out}}\times r}$ and $\mathbf{V}\in\mathbb{R}^{d_{\text{in}}\times r}$ have orthonormal columns, and $\boldsymbol{\Sigma}=\mathrm{diag}(\sigma_1,\dots,\sigma_r)$ with $\sigma_1\ge \dots \ge \sigma_r\ge 0$.
Under the $r$-slot convention in Section~\ref{sec:symmetry}, this is exactly the canonical object of the LoRA update, namely its rank-wise singular decomposition in Eq.~\eqref{eq:compact_svd}.

\paragraph{Proof of Proposition~\ref{prop:repr_invariance}.}
Let $(\mathbf{B},\mathbf{A})$ and $(\mathbf{B}',\mathbf{A}')$ be related by a LoRA reparameterisation as defined in Definition~\ref{def:lora_symm_appendix}. Then
\[
\mathbf{B}'\mathbf{A}' = (\mathbf{B}\mathbf{G})(\mathbf{G}^{-1}\mathbf{A}) = \mathbf{B}\mathbf{A},
\]
so the two parameterisations induce the same matrix $\Delta \mathbf{W}$.
Since the SVD depends only on $\Delta \mathbf{W}$ itself, both parameterisations induce the same rank-wise singular decomposition up to the intrinsic ambiguities of the SVD (sign flips and rotations within degenerate singular subspaces).
Under the fixed deterministic post-processing used by W2T, the resulting representation is consistent up to floating-point precision.

\subsection{QR-Based Construction of the Canonical Object}
\label{sec:qr_svd_prove}

\paragraph{QR-based construction.}
We prove that the QR-based reduction applied to LoRA factors $(\mathbf{B},\mathbf{A})$ recovers exactly the same canonical rank-wise decomposition as direct SVD on the induced update matrix $\Delta \mathbf{W} = \mathbf{B}\mathbf{A}$.

\paragraph{Proof of Proposition~\ref{prop:qr_svd}.}
Let $\Delta \mathbf{W} = \mathbf{B}\mathbf{A}$ with
$\mathbf{B} \in \mathbb{R}^{d_{\text{out}}\times r}$ and
$\mathbf{A} \in \mathbb{R}^{r\times d_{\text{in}}}$.
Let thin QR decompositions be
\begin{equation}
\mathbf{B} = \mathbf{Q}_B \mathbf{R}_B,\qquad \mathbf{A}^\top = \mathbf{Q}_A \mathbf{R}_A,
\end{equation}
where $\mathbf{Q}_B\in\mathbb{R}^{d_{\text{out}}\times r}$ and $\mathbf{Q}_A\in\mathbb{R}^{d_{\text{in}}\times r}$ have orthonormal columns
($\mathbf{Q}_B^\top \mathbf{Q}_B = \mathbf{I}_r$, $\mathbf{Q}_A^\top \mathbf{Q}_A = \mathbf{I}_r$),
and $\mathbf{R}_B,\mathbf{R}_A\in\mathbb{R}^{r\times r}$ are upper triangular.
Define the core matrix
\begin{equation}
\mathbf{M} := \mathbf{R}_B \mathbf{R}_A^\top \in \mathbb{R}^{r\times r},
\end{equation}
and let its SVD be
\begin{equation}
\mathbf{M} = \hat{\mathbf{U}} \boldsymbol{\Sigma} \hat{\mathbf{V}}^\top,
\end{equation}
with $\boldsymbol{\Sigma} = \mathrm{diag}(\sigma_1,\dots,\sigma_r)$ non-negative and sorted.
Then, defining
\begin{equation}
\mathbf{U} := \mathbf{Q}_B \hat{\mathbf{U}},\qquad \mathbf{V} := \mathbf{Q}_A \hat{\mathbf{V}},
\end{equation}
we have
\begin{equation}
\Delta \mathbf{W} = \mathbf{U} \boldsymbol{\Sigma} \mathbf{V}^\top,
\end{equation}
and $(\mathbf{U},\boldsymbol{\Sigma},\mathbf{V})$ forms an exact SVD of $\Delta \mathbf{W}$.
We proceed in two steps.

\paragraph{Step 1: Reduction to an $r\times r$ core.}
Substituting the QR decompositions into $\Delta \mathbf{W}=\mathbf{B}\mathbf{A}$ gives
\begin{align}
\Delta \mathbf{W}
&= \mathbf{B}\mathbf{A}
= (\mathbf{Q}_B \mathbf{R}_B)\,\mathbf{A}
= (\mathbf{Q}_B \mathbf{R}_B)\,(\mathbf{R}_A^\top \mathbf{Q}_A^\top) \\
&= \mathbf{Q}_B (\mathbf{R}_B \mathbf{R}_A^\top) \mathbf{Q}_A^\top
= \mathbf{Q}_B \mathbf{M} \mathbf{Q}_A^\top.
\label{eq:dw_qmqt}
\end{align}
Thus $\Delta \mathbf{W}$ is obtained from $\mathbf{M}$ by left and right multiplication with column-orthonormal matrices $\mathbf{Q}_B$ and $\mathbf{Q}_A$.

\paragraph{Step 2: Singular values coincide and SVD factors lift exactly.}

\emph{(2a) Singular values: $\Delta \mathbf{W}$ and $\mathbf{M}$ share the same non-zero singular values.}
Using~\eqref{eq:dw_qmqt},
\begin{align}
\Delta \mathbf{W}^\top \Delta \mathbf{W}
&= (\mathbf{Q}_B \mathbf{M} \mathbf{Q}_A^\top)^\top (\mathbf{Q}_B \mathbf{M} \mathbf{Q}_A^\top) \\
&= \mathbf{Q}_A\, \mathbf{M}^\top (\mathbf{Q}_B^\top \mathbf{Q}_B)\, \mathbf{M}\, \mathbf{Q}_A^\top \\
&= \mathbf{Q}_A\, (\mathbf{M}^\top \mathbf{M})\, \mathbf{Q}_A^\top,
\label{eq:gram_relation}
\end{align}
where we used $\mathbf{Q}_B^\top \mathbf{Q}_B = \mathbf{I}_r$.
Equation~\eqref{eq:gram_relation} shows that $\Delta \mathbf{W}^\top\Delta \mathbf{W}$ is an orthogonal (more precisely, column-orthonormal) congruence of $\mathbf{M}^\top \mathbf{M}$.
Therefore, $\Delta \mathbf{W}^\top \Delta \mathbf{W}$ and $\mathbf{M}^\top \mathbf{M}$ have the same non-zero eigenvalues.
Since the singular values are the square roots of the eigenvalues of these Gram matrices, $\Delta \mathbf{W}$ and $\mathbf{M}$ share the same non-zero singular values.
Consequently, the diagonal entries of $\boldsymbol{\Sigma}$ obtained from the SVD of $\mathbf{M}$ are exactly the singular values of $\Delta \mathbf{W}$, under the same sorting convention.

\emph{(2b) Lifting singular vectors: constructing an exact SVD of $\Delta \mathbf{W}$.}
From the SVD of $\mathbf{M}$, we have $\mathbf{M} = \hat{\mathbf{U}} \boldsymbol{\Sigma} \hat{\mathbf{V}}^\top$.
Substituting into~\eqref{eq:dw_qmqt} gives
\begin{equation}
\Delta \mathbf{W}
= \mathbf{Q}_B (\hat{\mathbf{U}} \boldsymbol{\Sigma} \hat{\mathbf{V}}^\top) \mathbf{Q}_A^\top
= (\mathbf{Q}_B \hat{\mathbf{U}})\, \boldsymbol{\Sigma}\, (\mathbf{Q}_A \hat{\mathbf{V}})^\top.
\end{equation}
Define $\mathbf{U} := \mathbf{Q}_B \hat{\mathbf{U}}$ and $\mathbf{V} := \mathbf{Q}_A \hat{\mathbf{V}}$.
We now verify the SVD orthonormality conditions:
\begin{align}
\mathbf{U}^\top \mathbf{U}
&= \hat{\mathbf{U}}^\top (\mathbf{Q}_B^\top \mathbf{Q}_B)\hat{\mathbf{U}}
= \hat{\mathbf{U}}^\top \mathbf{I}_r \hat{\mathbf{U}}
= \mathbf{I}_r,\\
\mathbf{V}^\top \mathbf{V}
&= \hat{\mathbf{V}}^\top (\mathbf{Q}_A^\top \mathbf{Q}_A)\hat{\mathbf{V}}
= \hat{\mathbf{V}}^\top \mathbf{I}_r \hat{\mathbf{V}}
= \mathbf{I}_r,
\end{align}
where we used $\mathbf{Q}_B^\top \mathbf{Q}_B = \mathbf{I}_r$, $\mathbf{Q}_A^\top \mathbf{Q}_A = \mathbf{I}_r$, and $\hat{\mathbf{U}}^\top\hat{\mathbf{U}} = \mathbf{I}_r$, $\hat{\mathbf{V}}^\top\hat{\mathbf{V}} = \mathbf{I}_r$.
Finally, $\boldsymbol{\Sigma}$ is diagonal with non-negative entries by definition of the SVD of $\mathbf{M}$.
Hence $(\mathbf{U},\boldsymbol{\Sigma},\mathbf{V})$ satisfies the defining conditions of a singular value decomposition of $\Delta \mathbf{W}$:
\begin{align}
\Delta \mathbf{W} &= \mathbf{U}\boldsymbol{\Sigma} \mathbf{V}^\top, \\
\mathbf{U}^\top \mathbf{U} &= \mathbf{I},\quad \mathbf{V}^\top \mathbf{V} = \mathbf{I}, \\
\boldsymbol{\Sigma} &= \mathrm{diag}(\sigma_1,\dots,\sigma_r),\quad \sigma_i \ge 0.
\end{align}

\paragraph{Remark (Equality to direct-SVD outputs).}
Under a fixed deterministic implementation convention (same routine/backend, singular-value ordering, and sign normalisation), the QR-based output numerically matches direct SVD on the same $\Delta \mathbf{W}$ up to floating-point precision.

% \subsection{Properties of the Singular-Triplet Representation}
% The Singular-Triplet Representation $\mathcal{S}(B,A)$ satisfies three key properties
% that justify its use as the input abstraction for Weight-to-Token modeling.
% \paragraph{(i) Symmetry resolution.}
% Because $\mathcal{S}$ is computed solely from $\Delta W$,
% it is invariant to the $\mathrm{GL}(r)$ reparameterization of LoRA factors.
% All factor pairs $(B,A)$ that induce the same update matrix
% map to the same representation up to intrinsic SVD ambiguities; under the fixed deterministic implementation convention used by W2T, this numerical representation is consistent up to floating-point precision.
% Thus, $\mathcal{S}$ provides a well-defined representation of LoRA updates in weight space.
% \paragraph{(ii) Lossless representation.}
% Since $\Delta W = BA$ with $B \in \mathbb{R}^{d_{\text{out}}\times r}$
% and $A \in \mathbb{R}^{r\times d_{\text{in}}}$,
% its rank is at most $r$.
% The singular value decomposition therefore captures the update exactly,
% without truncation or approximation.
% No information contained in $\Delta W$ is discarded.
% \paragraph{(iii) Rank-structured decomposition.}
% The representation expresses $\Delta W$
% as a sum of rank-1 components \citep{eckart1936approximation}, with at most $r$
% nonzero singular components under LoRA rank constraints.
% This rank-structured form naturally aligns with token-based modeling,
% where each singular component can be treated as an atomic weight token.

\begin{table*}[t]
\centering
\caption{Dataset-level summary of LoRA construction in Appendix~B. Shared backbone choices are described in the surrounding text.}
\label{tab:lora_dataset_summary}
\maintablesize
\setlength{\tabcolsep}{3.6pt}
\renewcommand{\arraystretch}{1.07}

\begin{tabular*}{\textwidth}{@{\extracolsep{\fill}}
    >{\raggedright\arraybackslash}m{2cm}
    >{\raggedright\arraybackslash}m{3.5cm}
    >{\raggedright\arraybackslash}m{6cm}
    >{\centering\arraybackslash}m{1cm}
    >{\raggedright\arraybackslash}m{2.5cm}
    @{}}
\toprule
Dataset & Raw data $\rightarrow$ LoRA & Key settings & Size & Checkpoint label \\
\midrule
CelebA & one identity $\rightarrow$ one LoRA; 21 images per identity & q/v; $r{=}8$; $\alpha{=}27$; lr 1e-4/3e-4/1e-3/3e-3; steps 100/133/167/200 & 10,177 & 40-d identity-level attribute vector \\
\midrule
CUB & one image $\rightarrow$ one LoRA & q/v; $r{=}8$; $\alpha{=}27$; lr 1e-4/3e-4/1e-3/3e-3; steps 100/150/200/250 & 11,788 & 312-d image-level attribute vector \\
\midrule
GoEmotions & one train sample $\rightarrow$ one LoRA & q/v; $r{=}8$; $\alpha{=}16$; lr 5e-6--3e-4; epoch 1--4; dropout 0--0.1 & 20,000 & 28-d emotion multi-hot vector \\
\midrule
ARC-Easy & one full-data trial $\rightarrow$ one LoRA & q/v; $r{=}8$; $\alpha{=}16$; lr 5e-6--3e-4; epoch 1--4; dropout 0--0.1 & 10,000 & checkpoint accuracy \\
\midrule
\multirow{2}{3.0cm}{\raggedright ARC-Challenge \\ BoolQ \\ GSM8K \\ MBPP} & Full-datasets runs $\rightarrow$ gallery LoRAs & q/v; $r{=}8$; $\alpha{=}16$; lr 5e-6--3e-4; epoch 1--4; dropout 0--0.1 & 1,296 & task identity \\
 & few-shot runs $\rightarrow$ query LoRAs & q/v; $r{=}8$; $\alpha{=}16$; lr 2e-4; epoch 3; dropout 0.05; shots 1/8/16/64/128/256; 25 queries per shot & 600 & task identity \\
\bottomrule
\end{tabular*}
\end{table*}

\section{LoRA Dataset Preparation}
\label{sec:appendix_data_prep}
\noindent Each benchmark item in our experiments is a complete LoRA checkpoint. The label is therefore attached to the checkpoint rather than to an individual image, token, or prompt. For attribute datasets, the label records semantic attributes inherited from the data used to train that adapter; for performance prediction, it records held-out downstream evaluation scores; for retrieval, it records the task identity used to define query--gallery relevance. Unless otherwise noted, downstream models operate on fixed 8:1:1 train/validation/test partitions over checkpoint names so that all encoders use identical splits.

\subsection{Vision Tasks}

\paragraph{Base model, adapter injection, and training randomization.}
Table~\ref{tab:lora_dataset_summary} summarizes, for each collection, how raw examples are converted into LoRA checkpoints, which key training settings vary across runs, and what checkpoint-level label is attached afterward.
On the vision side, all main adapters use Stable Diffusion v1.4 with LoRA on the U-Net query/value projections, $512{\times}512$ resolution, batch size $1$, constant learning rate, and no warmup.
The transfer settings in the main text reuse the same pipeline with either SD1.4 or SD1.5.
Checkpoint-level label construction is described separately for CelebA-LoRA and CUB-LoRA below.

\paragraph{CelebA-LoRA.}
For CelebA-LoRA, we first group images by identity using the CelebA identity metadata, then create one LoRA concept per identity.
For each identity, 21 images are selected; if an identity has fewer than 21 images, sampling with replacement is used and repeated images are materialized with suffixed filenames for deterministic bookkeeping.
The label of an adapter is the identity-level attribute vector: image-level CelebA annotations are averaged over the selected images and thresholded at zero, so each of the 40 attributes records the majority sign within that identity-specific training set.
This yields 10,177 LoRA samples, each supervised only by its checkpoint-level 40-dimensional binary label.

\paragraph{CUB-LoRA.}
CUB-LoRA uses one image per LoRA run, so the checkpoint label is directly tied to the source image.
Each run is launched from a temporary single-image folder, and the run name explicitly stores the CUB image ID for deterministic alignment after training.
Attribute labels are built from the official annotations and converted into a per-image binary table.
The resulting collection contains 11,788 LoRA checkpoints, each paired with 312 binary attributes.

\subsection{Language Tasks}

\paragraph{Base model and adapter injection.}
All language-side adapters use Llama-3.2-3B, apply LoRA to the attention query/value projections, and follow a shared 512-token context limit. Table~\ref{tab:lora_dataset_summary} summarizes the collection-specific raw-data mapping, compact training recipe, and checkpoint target used for language classification, performance prediction, and retrieval. As on the vision side, each checkpoint is stored together with its sampled hyperparameters and checkpoint-level supervision.

\paragraph{GoEmotions-LoRA.}
GoEmotions-LoRA trains one adapter per sampled training text.
The builder shuffles the train split, keeps 20,000 examples, and stores each checkpoint with its sorted 28-dimensional multi-hot emotion vector.

\paragraph{ARC-Easy-LoRA.}
For performance prediction, ARC-Easy-LoRA is built from 10,000 full-data training trials on ARC-Easy under randomized hyperparameter plans.
Each resulting adapter is evaluated on the task test split and stored together with the checkpoint-level targets accuracy.

\begin{table*}[!t]
    \centering
    \caption{Selected CelebA attributes (\%) where W2T is strongest or near-best against image-based methods. Gain is computed against the best image baseline in each row.}
    \label{tab:image_sota_comp}
    \maintablesize
    \setlength{\tabcolsep}{3.2pt}
    \renewcommand{\arraystretch}{1.05}
    \begin{tabular*}{\textwidth}{@{\extracolsep{\fill}}
        l
        >{\columncolor{black!8}}S[table-format=2.2]
        *{4}{S[table-format=2.2]}
        S[table-format=+1.2]
        @{}
    }
        \toprule
        \multirow{2}{*}{Attribute}
        & \multicolumn{1}{>{\columncolor{black!8}}c}{W2T}
        & \multicolumn{3}{c}{Image-based}
        & \multicolumn{2}{c}{Summary} \\
        \cmidrule(lr){2-2}\cmidrule(lr){3-5}\cmidrule(l){6-7}
        & \multicolumn{1}{>{\columncolor{black!8}}c}{(ours)}
        & \multicolumn{1}{c}{DMM-CNN}
        & \multicolumn{1}{c}{PS-MCNN-LC}
        & \multicolumn{1}{c}{GNAS}
        & \multicolumn{1}{c}{Best img.}
        & \multicolumn{1}{c}{Gain} \\
        \midrule
        Wearing Necklace & \textbf{96.41} & 88.03 & 88.98 & 87.61 & 88.98 & \textbf{+7.43} \\
        Narrow Eyes      & \textbf{95.10} & 87.73 & 89.07 & 87.66 & 89.07 & \textbf{+6.03} \\
        Big Lips         & \textbf{76.47} & 72.93 & 73.13 & 71.79 & 73.13 & \textbf{+3.34} \\
        Straight Hair    & \textbf{88.24} & 84.72 & 85.96 & 84.77 & 85.96 & \textbf{+2.28} \\
        Rosy Cheeks      & \textbf{98.37} & 95.32 & 96.92 & 95.01 & 96.92 & \textbf{+1.45} \\
        Blurry           & \textbf{98.69} & 96.40 & 98.00 & 96.42 & 98.00 & \textbf{+0.69} \\
        Pale Skin        & \textbf{99.35} & 97.00 & 98.84 & 97.24 & 98.84 & \textbf{+0.51} \\
        \midrule
        Gray Hair        & 98.04 & 98.27 & \textbf{98.66} & 98.37 & 98.66 & -0.62 \\
        Male             & 98.04 & 98.29 & \textbf{98.81} & 98.50 & 98.81 & -0.77 \\
        \bottomrule
    \end{tabular*}
\end{table*}

\begin{table}[t]
\centering
\caption{Parameter counts of the representative CelebA models. Values are measured from the saved experiment checkpoints.}
\label{tab:celeba_impl}
\maintablesize
\setlength{\tabcolsep}{5.0pt}
\renewcommand{\arraystretch}{1.06}
\begin{tabular}{lccccc}
\toprule
 & \method & GLNet & MLP & CNN & ViT \\
\midrule
\rowcolor{black!8}Params (M) & 17.75 & 23.16 & 51.06 & 77.75 & 39.91 \\
\bottomrule
\end{tabular}
\end{table}

\paragraph{Retrieval pool and few-shot queries.}
The retrieval benchmark reuses the same LLM LoRA pipeline on four datasets: ARC-Challenge, BoolQ, GSM8K, and MBPP.
The gallery is formed from full-data adapters under this filtered pool, yielding 1,296 checkpoints in total (323 ARC-Challenge, 315 BoolQ, 328 GSM8K, and 330 MBPP).
The query set is generated from fixed-hyperparameter few-shot plans in which only the sampled training subset changes across runs, across six shot levels $\{1,8,16,64,128,256\}$.
After capping each dataset-shot cell to at most 25 queries, the final benchmark contains 600 query adapters (150 per dataset).
For retrieval, the checkpoint label is the downstream dataset name, and relevance is defined by shared task identity between query and gallery adapters.

\subsection{Representative Implementation Details}
\label{sec:impl_details}
To make the encoder setup more concrete, we summarize the representative CelebA implementations used in the main attribute-classification experiments in Table~\ref{tab:celeba_impl}.
For \method, each checkpoint is first converted into its canonical rank-wise decomposition and then processed by a hierarchical Transformer with hidden dimension 128, one rank-level encoder layer, two layer-level encoder layers, four attention heads, and a 64-width classification head. We train for 45 epochs with batch size 64 using AdamW, and weight decay $10^{-3}$. Training begins with a 4-epoch warmup stage that updates only the projector, embeddings, and classifier, followed by full-model optimization with cosine annealing.
For the baseline encoders, we reuse their encoder backbones under the same CelebA-LoRA split and attach matched multilabel prediction heads. GLNet uses one equivariant layer with width 64 and an invariant head of width 64. The MLP and CNN baselines use hidden width 64 with a 16-width prediction head, while the ViT baseline uses token size 2048, embedding dimension 1024, depth 3, and four attention heads. All baselines are trained for 45 epochs with batch size 100 using AdamW, learning rate $10^{-3}$, weight decay $10^{-3}$, and cosine annealing.
Algorithm~\ref{alg:w2t_pipeline} summarizes the task-agnostic forward path used by \method. In the actual implementation, \textsc{Tokenize} corresponds to directional projection of left/right singular vectors, fusion, and singular-value FiLM modulation, while \textsc{RankPool} denotes the weighted rank pooling module used after optional rank-level self-attention.

\begin{algorithm}[H]
    \caption{Canonical Weight-to-Token encoding and prediction.}
    \label{alg:w2t_pipeline}
    \maintablesize
    \begin{algorithmic}[1]
        \State \textbf{Input:} LoRA factors $\{(B_p, A_p)\}_{p \in \mathcal{P}}$ with layer/module metadata
        \For{each position $p \in \mathcal{P}$}
            \State Compute thin QR: $B_p = Q_B R_B$, $A_p^\top = Q_A R_A$
            \State Form $M_p \gets R_B R_A^\top$ and compute $M_p = \hat{U}\Sigma \hat{V}^\top$
            \State Lift singular vectors: $U_p \gets Q_B \hat{U}$, $V_p \gets Q_A \hat{V}$
            \For{each rank component $k$}
                \State $\tau_{p,k} \gets \mathrm{Tokenize}(u_{p,k}, v_{p,k}, \sigma_{p,k})$
            \EndFor
            \State $\tilde{\tau}_{p,1:r} \gets f_{\mathrm{rank}}(\tau_{p,1:r})$
            \State $T_p \gets \mathrm{RankPool}(\tilde{\tau}_{p,1:r}, \sigma_{p,1:r})$
            \State $\tilde{T}_p \gets T_p + e_{\mathrm{layer}}(\ell(p)) + e_{\mathrm{module}}(m(p))$
        \EndFor
        \State $G_{1:|\mathcal{P}|} \gets f_{\mathrm{pos}}(\tilde{T}_{1:|\mathcal{P}|})$
        \State $z \gets \mathrm{AttnPool}(G_{1:|\mathcal{P}|})$
        \State \Return $\hat{y} \gets g(z)$
    \end{algorithmic}
\end{algorithm}

\begin{figure*}[!t]
    \centering
    \subfigure[Gray Hair]{\includegraphics[width=0.24\textwidth]{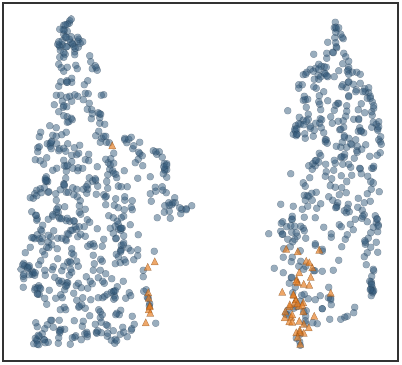}}
    \subfigure[High Cheekbones]{\includegraphics[width=0.24\textwidth]{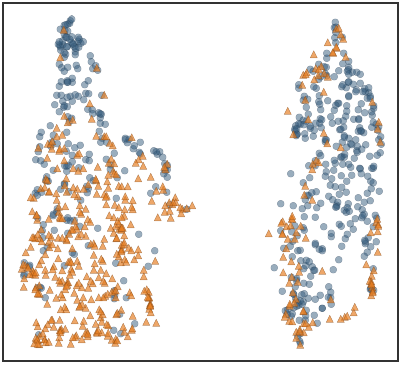}}
    \subfigure[Wearing Hat]{\includegraphics[width=0.24\textwidth]{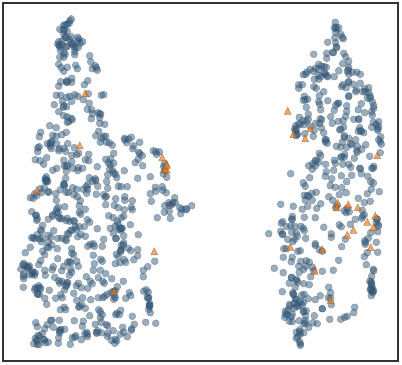}}
    \subfigure[Wearing Necklace]{\includegraphics[width=0.24\textwidth]{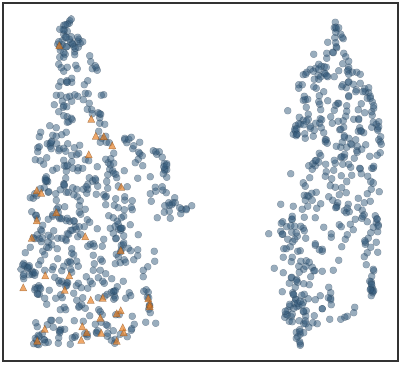}}
    \caption{Two-dimensional visualization of W2T weight-space embeddings for four highly separable CelebA attributes (\textit{Gray Hair}, \textit{High Cheekbones}, \textit{Wearing Hat}, and \textit{Wearing Necklace}). Each point represents one LoRA checkpoint and colors denote binary labels. Clear class-level clustering emerges across different attribute types, showing that LoRA weight space encodes semantically aligned structure.}
    \label{fig:extended_embedding}
\end{figure*}

\section{Additional Experiments}
\subsection{Attribute Classification}
% \paragraph{Scaling with Training Data}
% Figure \ref{fig:scaling} helps distinguish W2T from GLNet by showing how the two symmetry-aware approaches behave under different supervision budgets. In the low-data regime, GLNet performs better, suggesting that its stronger global equivariant inductive bias is advantageous when weight-space supervision is scarce. In this regime, W2T's richer architecture is not yet fully utilized.
% As the amount of training data increases, W2T improves more rapidly and eventually surpasses GLNet. This indicates that symmetry awareness is necessary, but not sufficient: once enough supervision is available, explicit canonical rank-wise components together with hierarchical composition become the more scalable representation strategy. In this sense, Figure \ref{fig:scaling} clarifies why W2T is not redundant given GLNet: GLNet is stronger in the extremely low-data regime, whereas W2T offers higher usable capacity once weight space supervision becomes sufficient.

\paragraph{Contextual comparison to image-based attribute recognition.}
\label{app:image_sota_comparison}
We include a small contextual comparison to image-based CelebA attribute recognition in order to illustrate what kind of semantic discrimination can be present in LoRA weight space. This is not a direct methodological comparison: image-based models observe raw pixels and are specifically designed for visual recognition, whereas W2T operates only on trained LoRA checkpoints.

Table \ref{tab:image_sota_comp} reports a small subset of representative attributes together with several image-based reference models. The purpose of this table is not to claim superiority over image-based recognition systems, but to contextualize that LoRA checkpoints trained from the same underlying dataset can still retain attribute-discriminative signal in weight space.

The comparison suggests that, for some attributes, directly modeling LoRA checkpoints as first-class data objects can recover surprisingly strong semantic distinctions even without pixel access at inference time. We therefore treat this table only as contextual evidence for the semantic content of LoRA weight space, rather than as a direct benchmark against image-based methods.

\paragraph{Embedding analysis.}
We provide a qualitative view of the learned W2T embedding space on several representative CelebA attributes. For each attribute, we extract the global weight-space embedding produced by W2T and visualize it in two dimensions using UMAP, where each point corresponds to one LoRA checkpoint and colors denote binary labels.

Figure \ref{fig:extended_embedding} shows that several attributes induce visible class-level structure in the learned embedding space. Separation is strongest for attributes associated with relatively global appearance changes, such as Gray Hair, Wearing Hat, and Wearing Necklace, while attributes requiring subtler local cues, such as High Cheekbones, remain only moderately separable.

These plots should be interpreted as qualitative evidence rather than a standalone proof. They are nevertheless consistent with the quantitative results in the main text, suggesting that W2T organizes LoRA checkpoints along semantically meaningful directions even when the model only observes weights and never accesses image pixels.

\subsection{Retrieval Analysis}
\label{app:retrieval_cluster_additional}
All retrieval results in this appendix follow the same setting as the main text. The encoder is trained on ARC-Easy performance prediction, while retrieval is evaluated on ARC-Challenge, BoolQ, GSM8K, and MBPP. The final pool contains 1,296 gallery adapters and 600 few-shot query adapters.
% We also compare against RawCos, which computes cosine similarity directly on raw LoRA weights, to isolate whether gains come from better representation geometry rather than from benchmark-specific preprocessing.

% \begin{table}[t]
%     \centering
%     \caption{W2T retrieval quality under varying query-shot budgets. Relative cost is normalized by the 256-shot setting.}
%     \label{tab:fewshot_shot_tradeoff}
%     \maintablesize
%     \setlength{\tabcolsep}{4.2pt}
%     \renewcommand{\arraystretch}{1.08}
%     \begin{tabular*}{\columnwidth}{@{\extracolsep{\fill}}l c r r@{}}
%         \toprule
%         Shots & Rel. cost & Hit@1 (\%) & NDCG@10 (\%) \\
%         \midrule
%         1   & $1/256$ & 36.00 & 38.72 \\
%         8   & $1/32$  & 50.40 & 51.39 \\
%         16  & $1/16$  & 58.93 & 54.26 \\
%         64  & $1/4$   & 58.41 & 57.10 \\
%         128 & $1/2$   & 56.00 & 52.70 \\
%         256 & $1$     & 61.95 & 61.56 \\
%         \bottomrule
%     \end{tabular*}
% \end{table}

\begin{table}[t]
    \centering
    \caption{W2T retrieval quality under varying query-shot budgets. Relative cost is normalized by the 256-shot setting.}
    \label{tab:fewshot_shot_tradeoff}
    \maintablesize
    \setlength{\tabcolsep}{4.2pt}
    \renewcommand{\arraystretch}{1.08}
    \begin{tabular*}{\columnwidth}{@{\extracolsep{\fill}}l c r r@{}}
        \toprule
        Shots & Rel. cost & Hit@1 (\%) & NDCG@10 (\%) \\
        \midrule
        1   & $1/256$ & 38.00 & 42.77 \\
        8   & $1/32$  & 63.00 & 60.94 \\
        16  & $1/16$  & 62.00 & 61.90 \\
        64  & $1/4$   & 69.00 & 65.71 \\
        128 & $1/2$   & 65.00 & 63.65 \\
        256 & $1$     & 71.00 & 72.03 \\
        \bottomrule
    \end{tabular*}
\end{table}

\begin{figure}[!tb]
    \centering
    \includegraphics[width=0.98\linewidth]{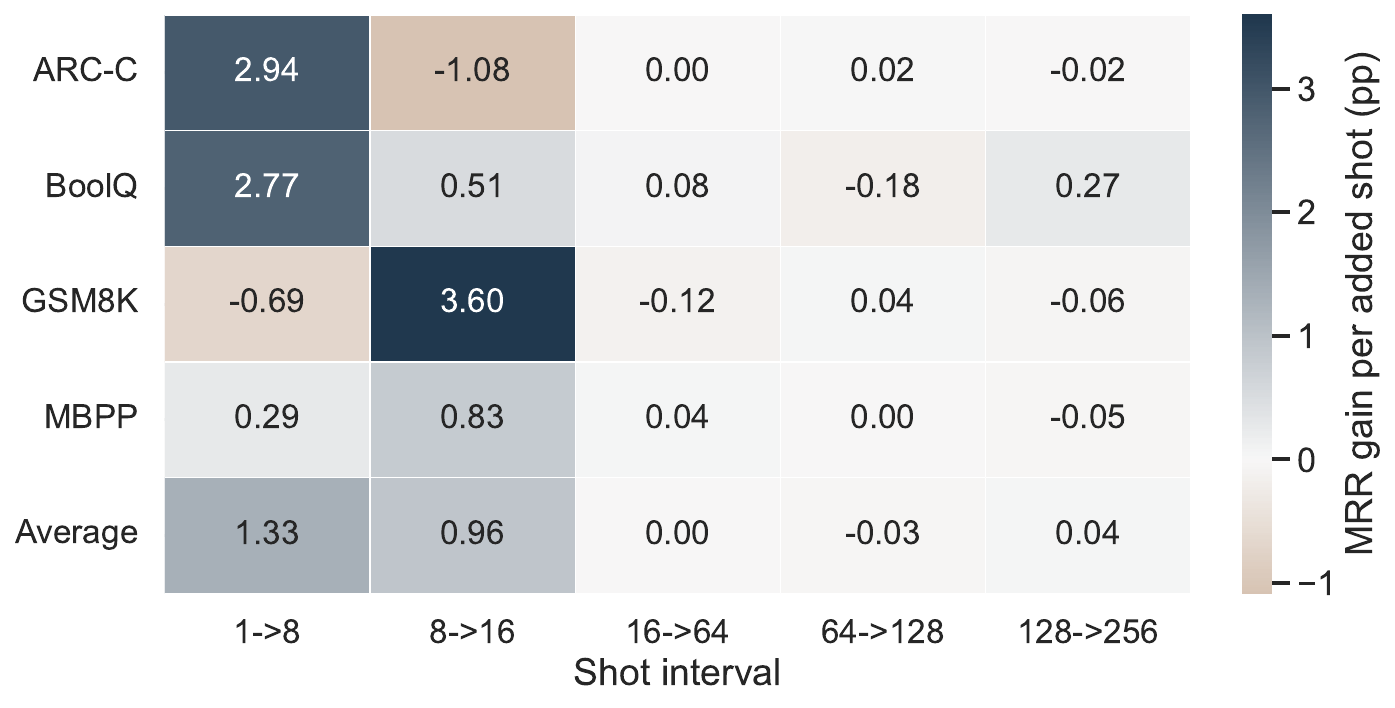}
    \caption{\textbf{Marginal MRR gain between successive query-shot levels under W2T retrieval.} Early budget increases produce the largest returns, while later increments yield smaller and more dataset-dependent improvements.}
    \label{fig:shot_scaling}
\end{figure}

Table \ref{tab:fewshot_shot_tradeoff} and Figure \ref{fig:shot_scaling} serve different but complementary roles. Table \ref{tab:fewshot_shot_tradeoff} reports the overall trade-off between retrieval quality and relative query cost at each shot level, while Figure \ref{fig:shot_scaling} highlights where additional shots still produce meaningful marginal gains. Taken together, they show that larger shot budgets do help, but the gain is highly nonlinear: most of the useful improvement is already obtained before the budget becomes large. In this sense, 8-shot and 16-shot remain favorable operating points, because they capture much of the achievable gain before the curve begins to flatten.

% Figure \ref{fig:fewshot_query_scatter} and its discussion have been moved to the main text (Section \ref{sec:fewshot_retrieval}).

Together, Table \ref{tab:fewshot_shot_tradeoff} and Figure \ref{fig:shot_scaling} suggest that the practical value of 8-shot and 16-shot adapter retrieval comes not only from lower cost, but also from the fact that symmetry-aware embeddings already induce a sufficiently task-consistent local neighborhood around a weak query adapter (see Figure \ref{fig:fewshot_query_scatter} in the main text).

% \paragraph{Complete retrieval matrix.}
% For completeness, Table~\ref{tab:fewshot_retrieval_fullmatrix} reports the full matrix of retrieval metrics under the same no-ARC-Easy mixed-pool setting, grouped by query-shot level. Each block includes one aggregated row and all per-dataset rows, with all methods shown side by side.

% \input{table/fewshot_retrieval_fullmatrix}

% \paragraph{Embedding retrieval and clustering.}
% Figure \ref{fig:retrieval_cluster_appendix} summarizes three selected
% CelebA retrieval/clustering metrics and visualizes relative improvement
% against GLNet.
% The largest gain is on Jaccard-gap, indicating stronger inter-cluster separation.
% W2T also maintains gains on Recall@10 and intra-Jaccard, supporting both retrieval coverage and cluster compactness.
% With ViT included, we observe that non-structure-aware baselines can obtain moderate retrieval coverage but still fail to form compact and well-separated semantic clusters.

% \begin{figure*}[htbp]
%     \centering
%     \subfigure[Task-aligned key metrics.]{
%         \includegraphics[width=0.58\textwidth]{figure/retrieval_cluster_selected.pdf}
%     }\hfill
%     \subfigure[Relative improvement over GLNet.]{
%         \includegraphics[width=0.36\textwidth]{figure/retrieval_cluster_gain_vs_glnet.pdf}
%     }
%     \caption{Additional retrieval and clustering visualizations on CelebA-LoRA embeddings.}
%     \label{fig:retrieval_cluster_appendix}
% \end{figure*}

\end{document}